\documentclass[%
 aip,
rsi,
 amsmath,amssymb,
 reprint,%
nofootinbib]{revtex4-1}
\usepackage{graphicx}
\usepackage{amsmath}
\usepackage[bb=libus]{mathalpha}
\usepackage[amssymb]{SIunits}
\draft 

\usepackage{color}
\usepackage[normalem]{ulem}

\usepackage{xcolor}

\begin{document}


\title{Multi-camera orientation tracking method for anisotropic particles in particle-laden flows} 

\author{Mees M. Flapper}

\affiliation{Physics of Fluids Department and Max Planck Center for Complex Fluid Dynamics, J. M. Burgers Centre for Fluid Dynamics, University of Twente, P.O. Box 217, 7522NB Enschede, The Netherlands}

\author{Elian Bernard}
\affiliation{ENS de Lyon, CNRS, LPENSL, UMR5672, 69342, Lyon cedex 07, France}

\author{Sander G. Huisman}
\affiliation{Physics of Fluids Department and Max Planck Center for Complex Fluid Dynamics, J. M. Burgers Centre for Fluid Dynamics, University of Twente, P.O. Box 217, 7522NB Enschede, The Netherlands}

\date{\today}

\begin{abstract}
    A method for particle orientation tracking is developed and demonstrated specifically for anisotropic particles. Using (high-speed) multi-camera recordings of anisotropic particles from different viewpoints, we reconstruct the 3D location and orientation of these particles using their known shape. This paper describes an algorithm which tracks the location and orientation of multiple anisotropic particles over time, enabling detailed investigations of location, orientation, and rotation statistics. The robustness and error of this method is quantified, and we explore the effects of noise, image size, the number of used cameras, and the camera arrangement by applying the algorithm to synthetic images. We showcase several use-cases of this method in several experiments (in both quiescent and turbulent fluids), demonstrating the effectiveness and broad applicability of the described tracking method. The proposed method is shown to work for widely different particle shapes, successfully tracks multiple particles simultaneously, and the method can distinguish between different types of particles.
\end{abstract}


\maketitle 

\section{Introduction}
\begin{figure}
    \centering
    \includegraphics[width=0.45\textwidth]{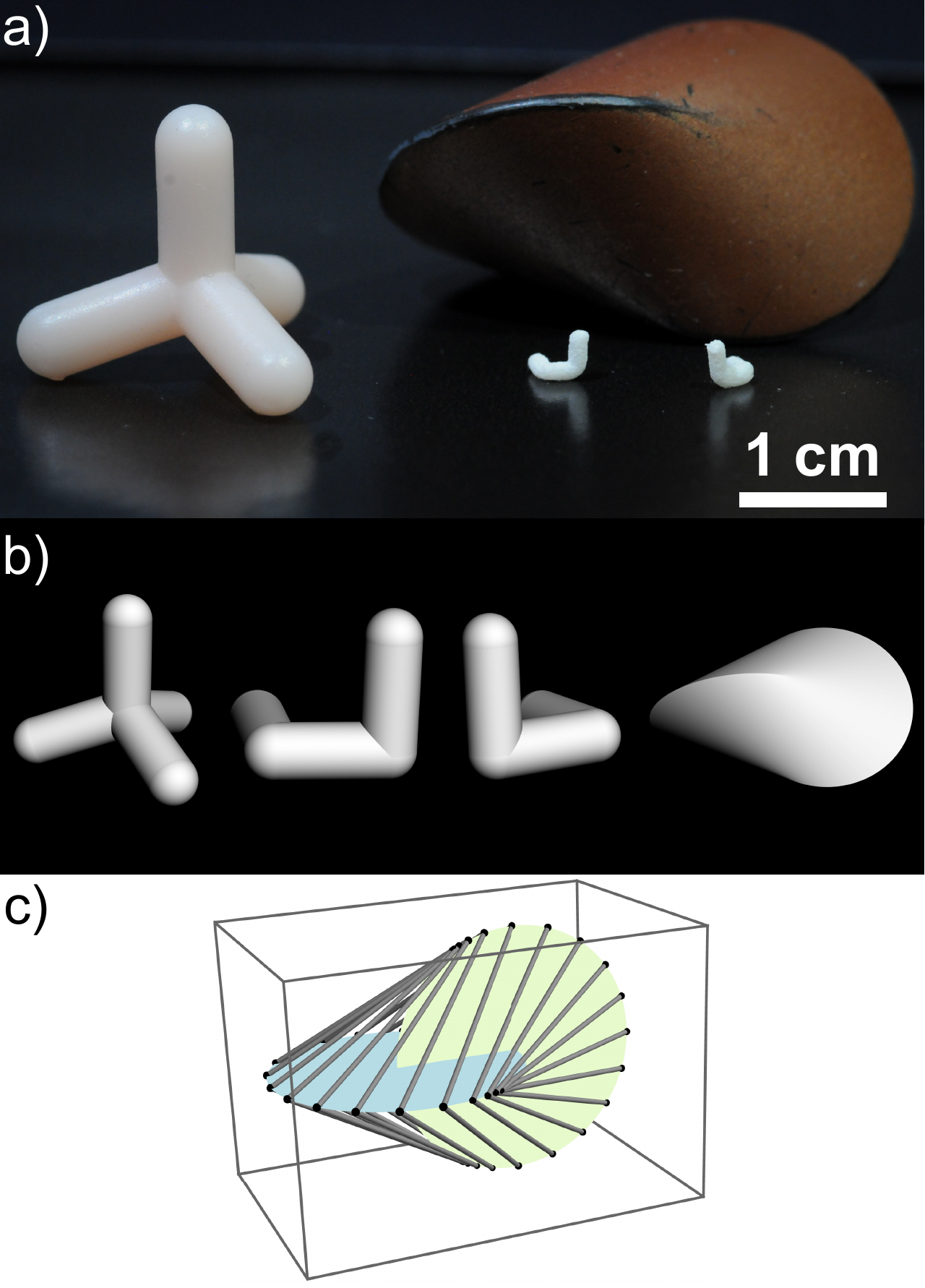}
    \caption{a) A photo of several 3D printed anisotropic particles tracked using the described method. From left to right, back to front: tetrad, oloid left-handed chiral particle, right-handed chiral particle. b) Synthetic 3D images of the used particles. From left to right: tetrad, left-handed chiral particle, right-handed chiral particle, oloid (not to scale). c) 3D sketch of the oloid geometry. The convex hull (`shrink-wrapping') of the two perpendicular disks gives the oloid surface. The grey lines span from one disk to the other, indicating the oloid surface.}
    \label{fig:Particles_photo}
\end{figure}

Tracking objects over time is vital for understanding dynamical systems and behaviours, ranging from pedestrian behaviour in crowds \cite{Corbetta2018}, sheep herd behaviour in a bottleneck \cite{Garcimartin2015}, to the transportation of particles on the water surface \cite{Salmon2023}. In fluid mechanics, particle tracking is a valuable instrument for investigating flows, particle-fluid interactions, and particle dynamics\cite{Schroder2022}. Tracking small tracer particles can be used to visualise complex dynamics in flows \cite{Godbersen2021}. Other fluid-dynamics applications of particle tracking are studying particle clustering dynamics \cite{Petersen2019}, investigating the settling of snowflakes \cite{Nemes2017}, or measuring the dynamics of complex particle geometries \cite{Collins2021}. To most effectively use particle tracking as a measurement tool, the particle tracking method should ideally be able to track multiple (types of) particles simultaneously, should be flexible in camera arrangement (the relative location and orientation of the cameras), and track both the location and orientation of the particles. Preferably the tracking method can track anisotropic particles, since particles in general are anisoptropic, and isotropic particles are the exception. Hitherto scientific research has mainly focused on isotropic (spherical) particles \cite{Brown2009,Chiarini2024}, though recently some have studied anisotropic particles \cite{Voth2017}. These particles range from rigid fibres \cite{Bakhuis2019} and spheroids \cite{Will2021}, to flexible fibres and discs \cite{Verhille2016,Verhille2022}, hexagonal platelets \cite{Tinklenberg2023}, and particles which break mirror symmetry \cite{Piumini2024}. Lagrangian tracking of these particles provides insights into particle dynamics, where both the particle's location and orientation are relevant for studying the flow-particle and particle-particle interactions\cite{Toschi2009,Brandt2022,Harms2024}. Many different tracking and measuring methods have been developed for obtaining the location and orientation data of spheres, fibres, and various anisotropic particles \cite{Cole2016,Mathai2016,Zimmermann2013,Ouellette2006,Ibarra2023}. In this work, we describe a tracking method which finds the location and orientation of anisotropic particles using multiple cameras to image the particles. This method is flexible in terms of the camera setup, and works for particles with concave shapes, even those which have a centre of mass outside of the particle. Previous works have demonstrated 3D orientation tracking methods based on matching multiple 2D orientations \cite{Parsa2012}, finding a particle's edges \cite{Brizzolara2021}, or a combination of the two \cite{Verhille2016}, where multiple particles can be tracked simultaneously \cite{Bounoua2018,Shaik2020,Baker2022}. More complex algorithms include methods using a particle marked with a black-and-white pattern \cite{Zimmermann2011,Mathai2016,Niggel2023} or projecting a synthetic particle model onto the experimental images \cite{Marcus2014,Cole2016}. This tracking method distinguishes itself from others by tracking multiple particle types simultaneously, being applicable to a wide range of particle shapes and sizes, including those that are concave and have their centre of mass outside the body. In addition, this method works for single or multiple cameras, and the camera arrangement is flexible. See Figure \ref{fig:Particles_photo}, the shown particles can be tracked using this method (including particles of convex shapes), and the robustness of the method is carefully examined, such that it can be used as a tool for accurate location and orientation tracking.
The working principle in this method is similar to previous methods using black-and-white particle projections, and relies on imaging a particle with one or multiple different directions. These images are then compared to calculated projections of a synthetic particle with a known orientation. The synthetic particle's orientation is varied until the correct orientation is found. The details of this method are described step-by-step in section \ref{sec: Tracking principles}. The robustness and accuracy of the method are quantified and evaluated in section \ref{sec: Robustness}, detailing the effects of noise and image size using synthetic data. To close, section \ref{sec:Application} shows the usage of this method in test experiments, where we tracked chiral particles (particles breaking mirror symmetry, giving a left-handed and right-handed particle), tetrads, and oloids, all shown in Figure \ref{fig:Particles_photo}.

\section{Tracking principles}
\label{sec: Tracking principles}
\begin{figure}
    \centering
    \includegraphics[width = 0.45\textwidth]{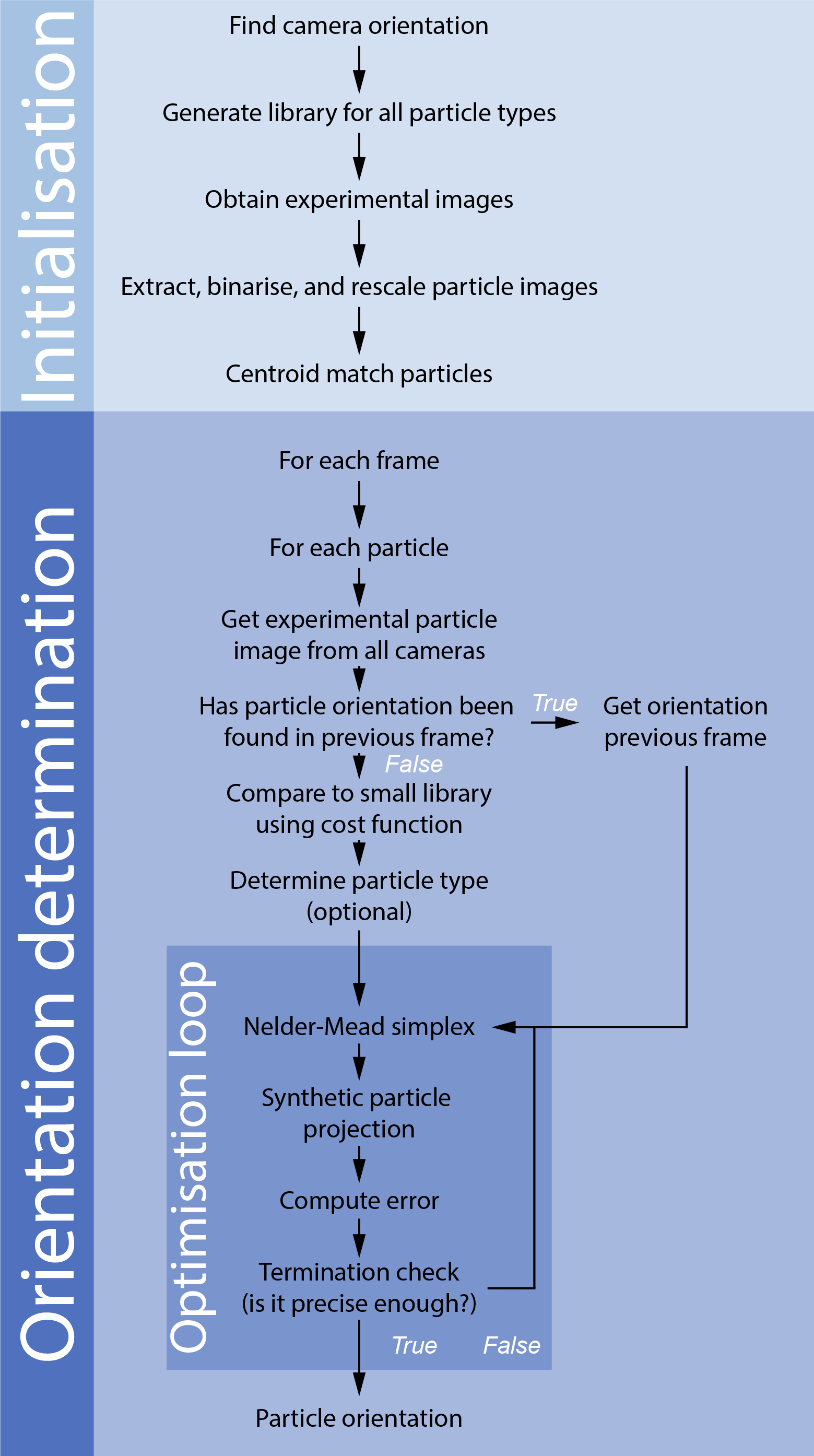}
    \caption{Illustrated workflow of the orientation tracking algorithm, describing the order of steps taken in the determination of the particle orientation.}
    \label{fig:Workflow}
\end{figure}

A wide range of particle tracking algorithms have been developed, with different strong and weak points. Methods based on finding the 2D orientation or detecting the particle endpoints in 2D are very effective for rod-like particles \cite{Parsa2012,Bakhuis2019,Brizzolara2021}, but are not suited for more complex 3D shapes. A more general approach for 3D particle tracking is carving voxels based on multiple camera images, then finding the particle orientation by fitting the particle to the voxels \cite{Alipour2021,Giurgiu2024}. This method is shown to work well for tracking, and allows for a wide variety of geometries or configurations. \cite{Fu2018,Masuk2019}. However, for our chiral particles, for specific camera configurations (for example when the cameras are closely arranged in a near-planar arrangement), we cannot distinguish the handedness of chiral the particle using this carving method. Therefore, for the chosen particle geometries, we do not have full camera flexibility when using the carving method. To have a flexible camera arrangement to track the anisotropic particles shown in Figure \ref{fig:Particles_photo}, we therefore defer to another method.\\
The basic principle in this method uses a synthetic particle with a known orientation, where the images of the synthetic particle are compared to the experimental images of a particle, similar to a method developed by Zimmermann et al \cite{Zimmermann2011}, and Mathai et al\cite{Mathai2016}. Zimmermann et al. developed an algorithm for tracking spheres, by creating a black-and-white pattern on the sphere. Synthetic images of the black-and-white pattern for many known orientations are stored in a database, covering all orientations in a course grid (grid spacing of approximately 12$^\circ$). They first sort the experimental images into groups of similar orientation (orientations differing by less than approximately 30$^\circ$--40$^\circ$) by comparing the experimental image to the course grid. The orientation of the particles is then refined using a fine grid, having a grid spacing of approximately 3$^\circ$, resulting in an orientation accuracy of around 3$^\circ$. This method by Zimmerman et al. was later improved by Mathai et al. \cite{Mathai2016}, who used a similar black-and-white pattern, although they did not use a library of known orientations, but rather used synthetically generated images only. The rotated and projected patterns are compared to the experimental images. By minimising a cost function of the experimental and synthetic images, an accuracy in orientation of $\mathcal{O}(0.1^{\circ})$ can be achieved. \\
Voth et al. use a different tracking method, also projecting a synthetic particle onto the experimental images. The synthetic particle is generating by modelling the light intensity distribution over the particle. The difference between the experimental images and synthetic projections are then minimised to find the orientation of the particle \cite{Marcus2014,Cole2016}.\\
We build on these techniques, but now utilise the silhouette of the particle to determine its orientation. We synthetically generate silhouettes of a particle with known orientation and compare to the experimentally recorded silhouette to determine the particle orientation. The method in this paper works in principle for any type of particle which is anisotropic to such a degree that the shape outline can be used to determine its orientation. The working principle can be used for any number of cameras in freely chosen camera arrangements. The materials needed to implement this method is therefore limited to high-speed cameras (two being the minimum for complex particles, and each extra camera improves the accuracy of the method) and a computer for processing the recorded images. We expand on previous literature by tracking multiple types of complex-shaped particles simultaneously, and by extensively quantifying the accuracy and robustness for this method, including effects of noise, image resolution and camera arrangements.\\
The developed orientation tracking algorithm consists of different steps, shown in the workflow in Figure \ref{fig:Workflow}. These steps are explained in detail in the following sections.

\subsection{Initialisation}
\begin{figure}
    \centering
    \includegraphics[width = 0.45\textwidth]{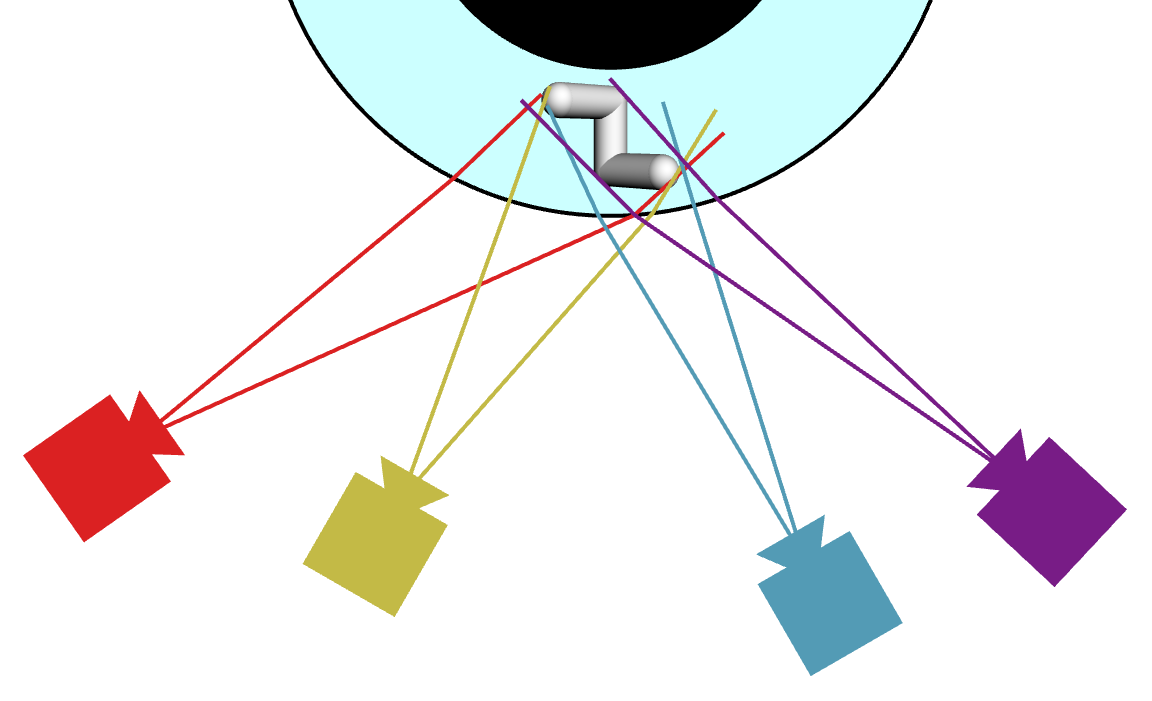}
    \caption{Top view illustration of cameras viewing a particle using camera rays (particle not to scale) inside a Taylor-Couette geometry. Note that the calibration accounts for diffraction at the air-outer cylinder, and outer cylinder-water interfaces.}
    \label{fig:Camera_setup_2D}
\end{figure}

\begin{figure}
    \centering
    \includegraphics[width = 0.45\textwidth]{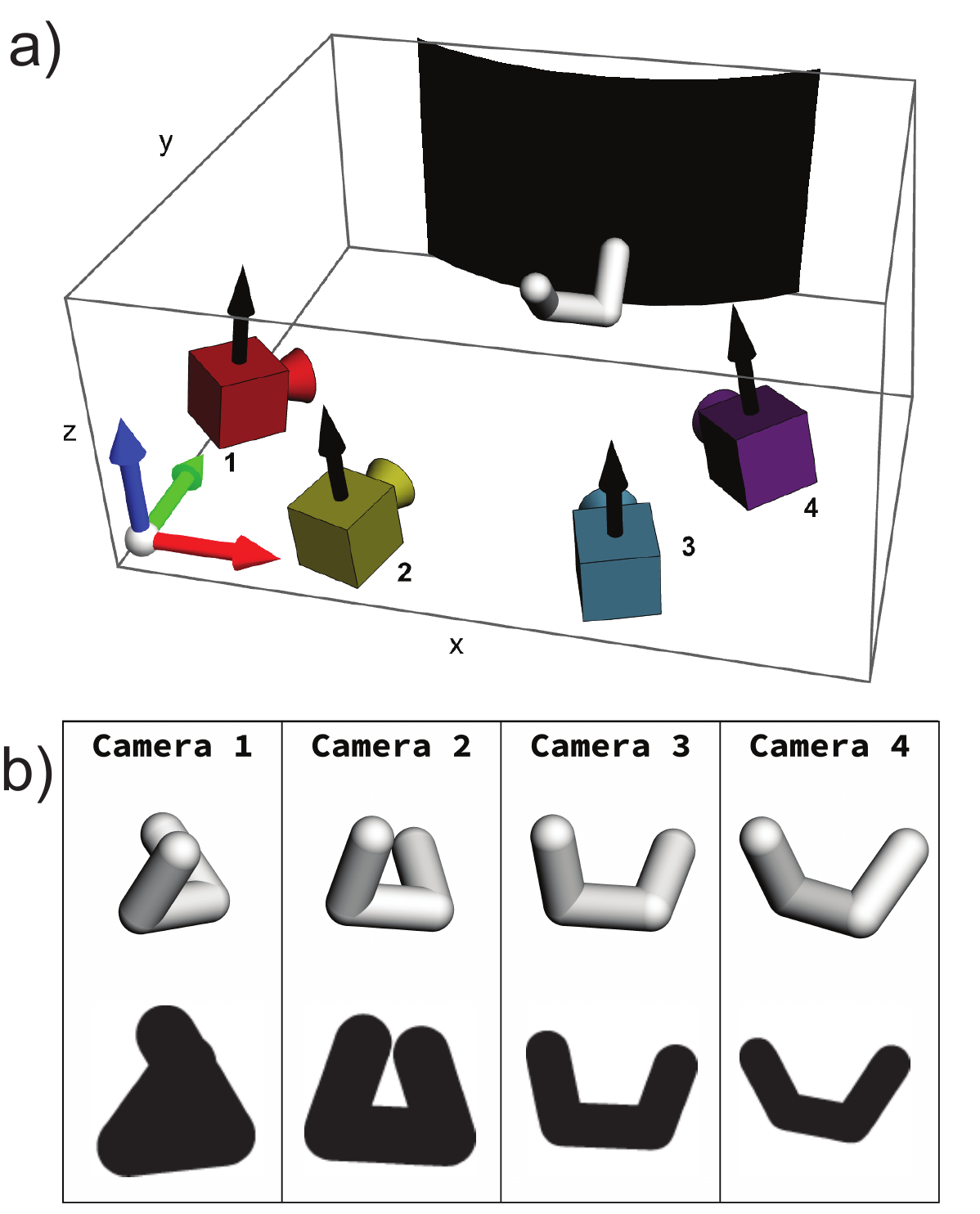}
    \caption{a) A 3D representation of the cameras viewing a chiral particle. The cameras are set up as in the Taylor--Couette geometry described in section \ref{sec:Application}. The curved black surface represents the inner cylinder of the Taylor--Couette. Graphic not to scale. b) The particle as viewed by the four cameras in 3D (shown in the top row), and its projections (shown in the bottom row).}
    \label{fig:particle_projections}
\end{figure}

\begin{figure}
    \centering
    \includegraphics[width=0.45\textwidth]{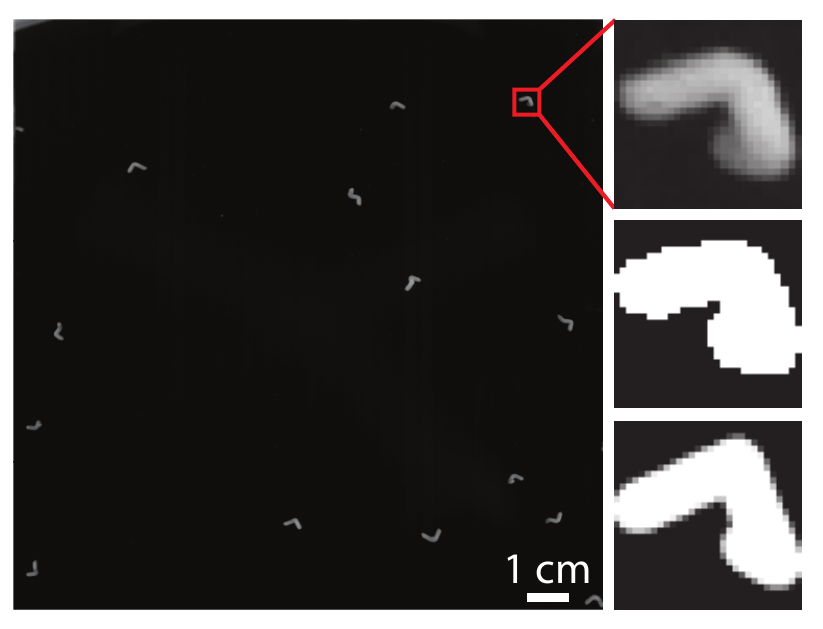}
    \caption{Basic steps of the tracking algorithm. The left figure shows the complete raw image for one of the cameras. The top-right figure shows an extracted particle from the raw image. The middle-right figure shows the binarised particle. The bottom-right figure shows a projection of the optimised orientation of this particle.}
    \label{fig:Tracking_steps}
\end{figure}

An important first step of the method is finding the camera location and orientation with respect to the lab frame and measurement volume, which can be done by various techniques to obtain the camera parameters for each camera \cite{Tsai1987}. The camera vectors differ for each location, as illustrated in Figure \ref{fig:Camera_setup_2D}. For our applications, we performed a calibration to find the camera locations and orientations based on a ray-matching algorithm by Bourgoin and Huisman \cite{Bourgoin2020}.
The next step is generating a library of standard orientations, similar to Zimmermann et al. and Mathai et al \cite{Zimmermann2011,Mathai2016}. This library is specific to the particle and the camera parameters, and will later be used to find a first guess for the particle orientation. Although this first guess could also be found by using a deep learning model to give an orientation as recently shown by di Giusto et al.\cite{diGiusto2024}, who used a neural network to determine the particle orientation based on images of particle projections. To generate the library, a standard orientation is defined first, where the vertices of the particle are defined as position vectors $\vec{v}$. Rotating the particle is done by rotating the vertices using quaternions. To write the particle as a quaternion, the particle vertex vectors are added with a zero: 
\begin{equation}
    \mathbf{p} = 0 + v_x \mathbb{i} + v_y \mathbb{j} + v_z \mathbb{k}.
\end{equation}
Here $\mathbb{i}, \mathbb{j}$ and $\mathbb{k}$ are imaginary axes. These quaternions of the particle vertices are then rotated into 1600 different standard orientations for our library. The used rotation quaternions are generated using 100 axes (equally distributed over a sphere) combined with 16 rotation angles (equally distributed between $-\pi$ and $\pi$). The generation of these rotation quaternions can be altered for particles with certain symmetries, like our chiral particles. As can be seen in Figure \ref{fig:Camera_setup_2D}, the chiral particle is symmetrical for $180^\circ$ rotations around its symmetry axis. Taking this symmetry into account prevents the generation of multiple equivalent orientations.\\
The rotation quaternion is defined as 
\begin{equation}
    \mathbf{q} = \cos{\left( \frac{\theta}{2} \right)} + (u_x \mathbb{i} + u_y \mathbb{j} + u_z \mathbb{k}) \sin{\left( \frac{\theta}{2} \right)},
\end{equation}
where $\theta$ is the rotation angle, and $\vec{u} = (u_x,u_y,u_z)$ is the axis of rotation (which has unit length). Here we note that the first term is real, whereas the second term is imaginary. This rotation quaternion is then used to rotate the particle from its standard orientation to a new orientation using
\begin{equation}
    \mathbf{p_r} = \mathbf{q p q}^{-1}.
\end{equation}
Here $\mathbf{p_r}$ is the rotated particle (in quaternion space), $\mathbf{p}$ is the quaternion describing the particle in its reference orientation, and $\mathbf{q}^{-1}$ is the reciprocal of $\mathbf{q}$.\\
The real part of the quaternion $\mathbf{p_r}$ gives the rotated particle vertices, which are used to display the particle as shown in Figure \ref{fig:particle_projections}.
Figure \ref{fig:particle_projections}a shows a synthetic particle imaged by four cameras (particle not to scale), where the cameras are set up as in a Taylor--Couette measurement discussed in Section \ref{sec:Application}. Figure \ref{fig:particle_projections}b shows the 3D particle as viewed by the cameras in the top row (using the previously found camera parameters). The bottom row shows the greyscale projections of the particles as viewed by the cameras. \\
For computational efficiency, the synthetic particle projections are generated by obtaining the projected vertices and subsequently drawing a line (with thickness corresponding to the particle thickness) from vertex to vertex. To obtain the synthetic particle projection as accurately as possible, the projection is initially generated in binary at four times the desired size, after which the projection is downsized to the desired size using a box-shaped interpolation. This makes the projections more accurate at the particle's edges, where the pixels at the edges are anti-aliased due to the interpolation, see Figure \ref{fig:particle_projections}b. The anti-aliasing techniques allow us to more accurately determine the orientation. \\
For each of the 1600 previously described rotation quaternions, we generate the synthetic projections of the particle for each camera and store these in the library, along with the rotation quaternion. A library is generated for each particle type used in the experiment (a left-handed and right-handed particle in the case of the used chiral particles).\\
Following the flowchart of Figure \ref{fig:Workflow}, the experimental images are recorded, after which the particles are isolated and extracted. We remove any small bubbles, as only particles should be extracted from the images. This can be achieved using a segmentation method of choice \cite{DigitalImageProcessing,Rother2004,DeBruijne2004} where necessary. Overlapping particles are cut out as a single image, but cannot be distinguished, and the overlapping particles' orientations cannot be easily found in the current state of the described algorithm. In theory, one could optimise the orientation of two particles simultaneously, but the large amount of local minima makes this a challenging endeavour. Once the particle images are extracted from the raw image, these images are processed, as shown in Figure \ref{fig:Tracking_steps}. The left panel shows the raw image from one of the cameras. The particles are extracted from the image as seen in the top-right image. The image is binarised as shown in the middle-right panel of Figure \ref{fig:Tracking_steps}.
The particles are centroid-matched between the multiple cameras using a method of choice. For our applications, the matching was performed using the previously mentioned ray-traversal method \cite{Bourgoin2020}. When this step is finished, we therefore have the 3D positions of the centroids of the particles, the cut-outs of the particles on each camera, and a library of synthetic particle projections in different orientations. Figure \ref{fig:Tracking_steps} shows the optimised orientation of the chiral particle in the bottom-right panel, which is found using the steps described below.

\subsection{Determining the orientation}
\begin{figure}
    \centering
    \includegraphics[width=0.45\textwidth]{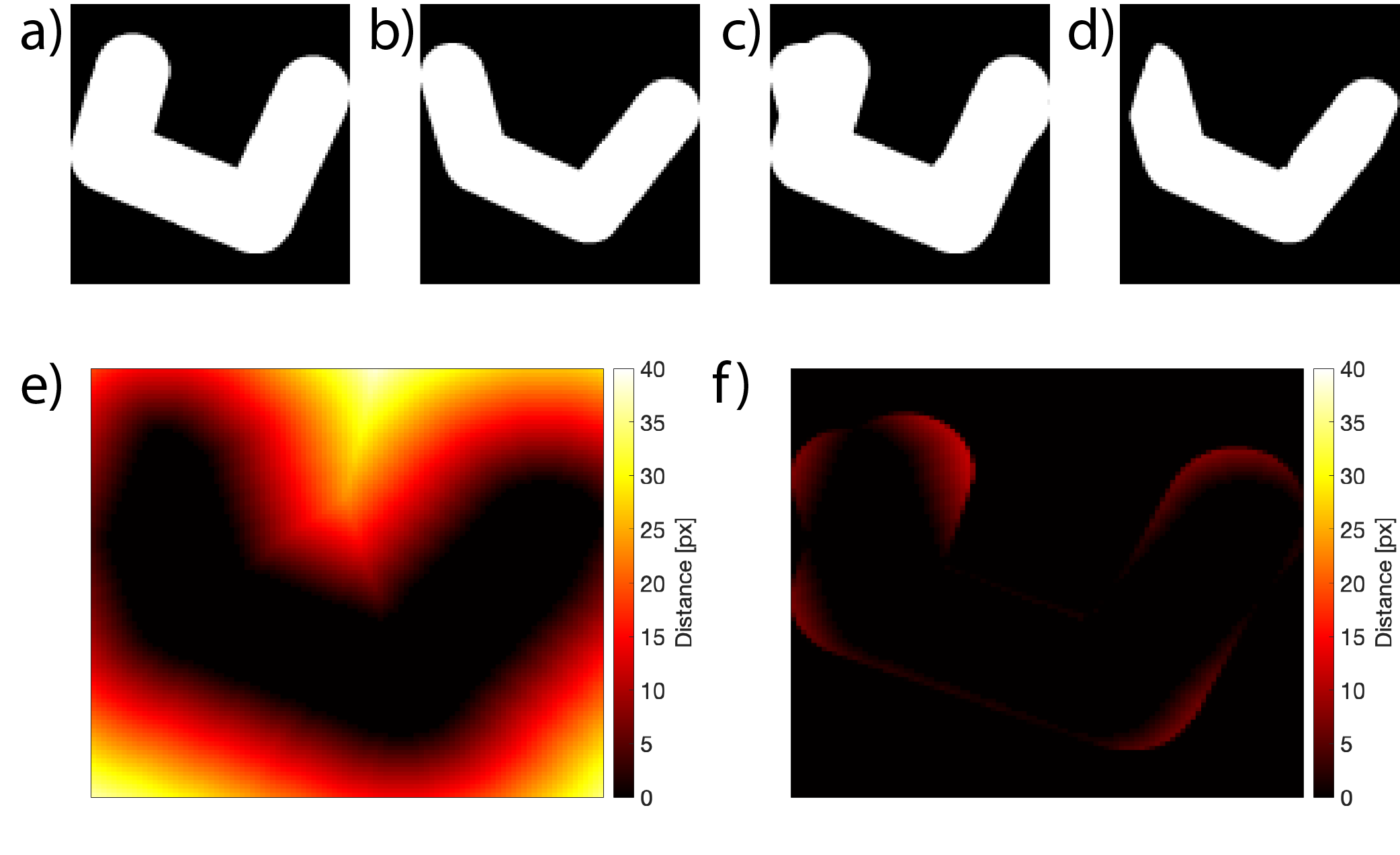}
    \caption{a) Example image 1 ($I_1$). b) Example image 2 ($I_2$). c) Union of images 1 and 2 ($U$). d) Intersection of images 1 and 2 ($O$). e) Distance transform ($D_{ij}$) of the overlap. f) Distance transform $D_{ij}$ multiplied by the union $U$. The sum of this matrix is normalised by the image width squared to give the error.}
    \label{fig:Error_function}
\end{figure}

The orientation is found by comparing the experimentally recorded images against the synthetic particle projections, following the workflow as shown in Figure \ref{fig:Workflow}. First a frame is selected, from which we select a single particle. Using the centroid-matching from before, we find the images of the selected particle on all cameras. Checking whether the selected particle's orientation has been found before determines how to make a first guess for the orientation. If this particle's orientation has been found before, use the previous orientation as a first guess for the current frame. For a particle which has not been found before, we compare the experimental images to the library images.\\
Comparing to the library of synthetic projections is done using a cost function, see Figure \ref{fig:Error_function}. 
The error is computed between the image and the library images by reshaping the experimental image and synthetic image to the same size (100 px $\times$ 100 px in this case) using MATLAB's imresize function with a box-shaped kernel for interpolation. Subsequently, we find the overlap (intersection) between both images, as shown in Figure \ref{fig:Error_function} c). The intersection is mathematically noted as
\begin{equation}
    O = I_1 \cap I_2.
\end{equation}
Panel (d) shows the union of the experimental and synthetic image, which is noted as
\begin{equation}
    U = I_1 \cup I_2.
\end{equation}
The distance transform of the overlap is computed in pixels and shown in Figure \ref{fig:Error_function}(e). This transform gives the minimum distance to a white pixel for each pixel of the overlap image, which is defined as
\begin{align}
    D_{kl} &= \min\left( \sqrt{(k - i)^2 +(l - j)^2} \right) \text{, with}\\
    (i,j) &\in O_{ij}|_{O_{ij}=1}.
\end{align}
The distance transform is multiplied by the union to find the mismatch between the two input images, as shown in panel (f). Here the pixels further from the experimental particle are penalised more strongly. We sum the found matrix and divide by the image dimensions to find the error, defined as
\begin{equation}
    \epsilon = \frac{U_{ij}  D_{ij}}{H W} ,
\end{equation}
where $W$ is the width of the image, and H the image height. This error is summed over the cameras to find the total error. This error is symmetrical, meaning the error value is the same when computing the error for image 1, using image 2 as a reference and vice versa.\\
The previously described error is computed for all library orientations, from which we select the 4 orientations with lowest error as first guesses for optimisation. In case multiple particle types are used, the errors of the 4 first guesses are compared to determine the particle type. The 4 different orientations (defined as the rotation quaternions) with lowest errors are each used to define a simplex for a Nelder--Mead minimisation\cite{Nelder1965}. Subsequently, these 4 different first guesses are all optimised using a Nelder--Mead algorithm by minimising the error. For each iteration in the Nelder--Mead algorithm, the simplex is given by five points (each describing an orientation), for which we generate synthetic particle images. The error is computed as before, and the Nelder--Mead simplex is updated accordingly until the loop is terminated (when the simplex hyper-volume is smaller than $10^{-8}$). The rotation quaternion with the lowest total error over the different cameras is selected as the orientation of the particle. A synthetically generated particle with the optimised orientation is shown in Figure \ref{fig:Tracking_steps} in the bottom-right panel.\\
As mentioned before, particles which overlap on one of the cameras could not be reliably distinguished and tracked while overlapping, which is a limitation in the current algorithm, and limits its use to cases of dilute particle concentrations. This limitation may be overcome by using the known particle shape for specific geometries \cite{Shen2000,Neoptolemou2022}, or a machine-learning aided approach may be used to better distinguish the overlapping particles \cite{Puzyrev2020}.

\subsection{Correcting the centre of mass}
The previously determined location of the detected particle was found using the centroid of the projections of the particle. In general, however, the centroid of the particle projection (the centroid of the 2D image) is different from the projection of the 3D centre of mass of the particle. 
\begin{figure}
    \centering
    \includegraphics[width= \columnwidth]{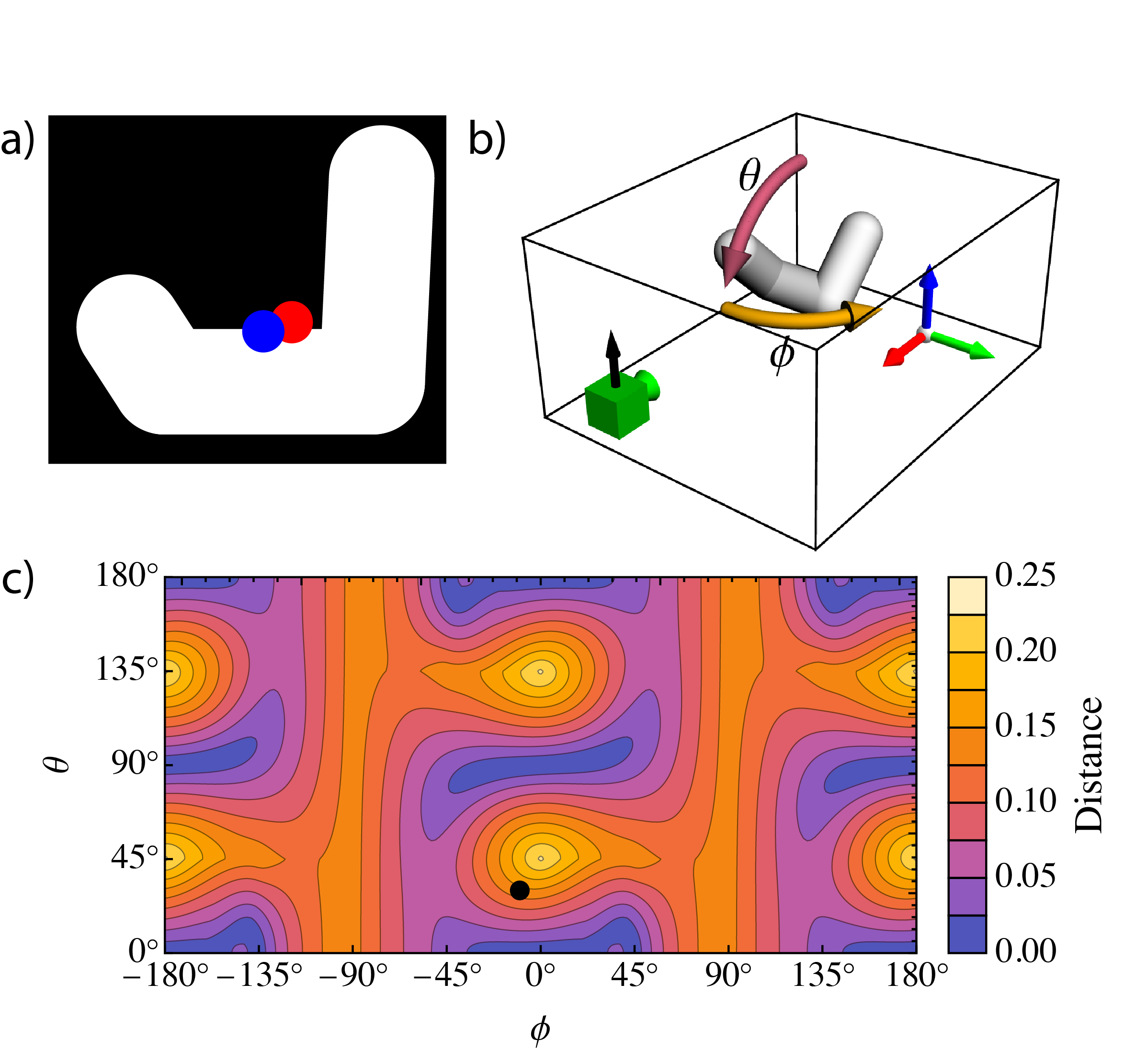}
    \caption{a) Projection of a chiral particle, with the centre of mass of the projection shown in red, the projection of the centre of mass shown in blue. b) Definitions of rotations ($\theta$ and $\phi$) to obtain particle orientations. The order of rotations is $\theta$, then $\phi$. A gnomon is included to indicate the Cartesian axes (red, green, blue denoting x, y, and z respectively). c) Distance (normalised by the particle arm length) between the centroid of a projected particle and the projected centre of mass, as a function of the particle orientation. The black point indicates the orientation of the particle shown in panel a.}
    \label{fig:Particle_coms}
\end{figure}
This point is illustrated in Figure \ref{fig:Particle_coms}a, showing a projection of a synthetic particle, where the centroid of the projection and the projection of the particle's centre of mass are indicated in red and blue respectively. The difference between the centroid of the projection and the projected COM has to be corrected to accurately find and track the centre of mass of the particle. As Figure \ref{fig:Particle_coms} shows, once the orientation of a particle has been found, the found orientation can be used to calculate the true centre of mass. We calculate the position of the centre of mass of the particle in the reference orientation, which is then rotated using the found rotation quaternion of the orientation. Generating the projections of this rotated synthetic particle (and its centre of mass) provides the projected coordinates of the centre of mass. The synthetic images are rescaled to the size of the experimental particle images to be able to compare the two points. Computing the difference between the centroid and the projected centre of mass for each camera then allows us to shift the previously found centroids to the corrected centre of mass coordinates. These corrected 2D centre of mass coordinates are then matched again to find the corrected 3D position of the centre of mass. \\
The magnitude of displacement from the centroid of the projected particle to the projection of the centre of mass depends on the particle orientation. Figure \ref{fig:Particle_coms}b shows the distance between the centroid of the projection and the projected centre of mass, normalised by the particle arm length.\\
If a high degree of accuracy is required, the corrected 3D position can be used to generate the camera vectors for the orientation optimisation, thereby iteratively optimising the position and orientation of the tracked particle which quickly converges. Alternatively, a simultaneous optimisation of the position and orientation could be implemented, as shown in other work \cite{Oehmke2021}.

\section{Robustness \& errors}
\label{sec: Robustness}
We can quantify the error in the orientation detection using synthetic data. The synthetic data consists of a large set of images of a particle with a known orientation, allowing us to quantify the errors and robustness of the method. The error in finding the orientation is defined by the angle between the found orientation quaternion and the known orientation quaternion of the synthetic particle. The angle between two quaternions is given by
\begin{equation}
    \theta_{\textrm{err}} = 2 \arcsin{\left(||\mathbf{q_{1,i}} \mathbf{q_{2,i}}^{-1}||\right)},
\end{equation}
where $q_{1,i}$ and $q_{2,i}$ are the imaginary parts of quaternions $1$ and $2$ respectively. Here the double bars denote taking the $\ell^2$ norm. Alternatively, the angle $\theta_{err}$ is given by $\theta_{\textrm{err}} = 2 \arccos \left (||\mathbf{q_{1,r}} \mathbf{q_{2,r}}^{-1}||\right)$, where we use the real part of the quaternions. The effect of noise, image size, number of cameras, and camera arrangement on the orientation error is tested by processing synthetic data.\\
This section only considers the error in the orientation, since this method focuses on finding the orientation. How the error in position affects the orientation and how the orientation error affects the corrected is shown in Appendix \ref{sec:Error propagation}. Furthermore, only the errors of the orientation of the chiral particles are investigated here. The results for all investigated particle geometries are comparable, as shown in Appendix \ref{sec:Error particles}.

\subsection{Noise}
\begin{figure*}
    \centering
    \includegraphics[width = 0.7\textwidth]{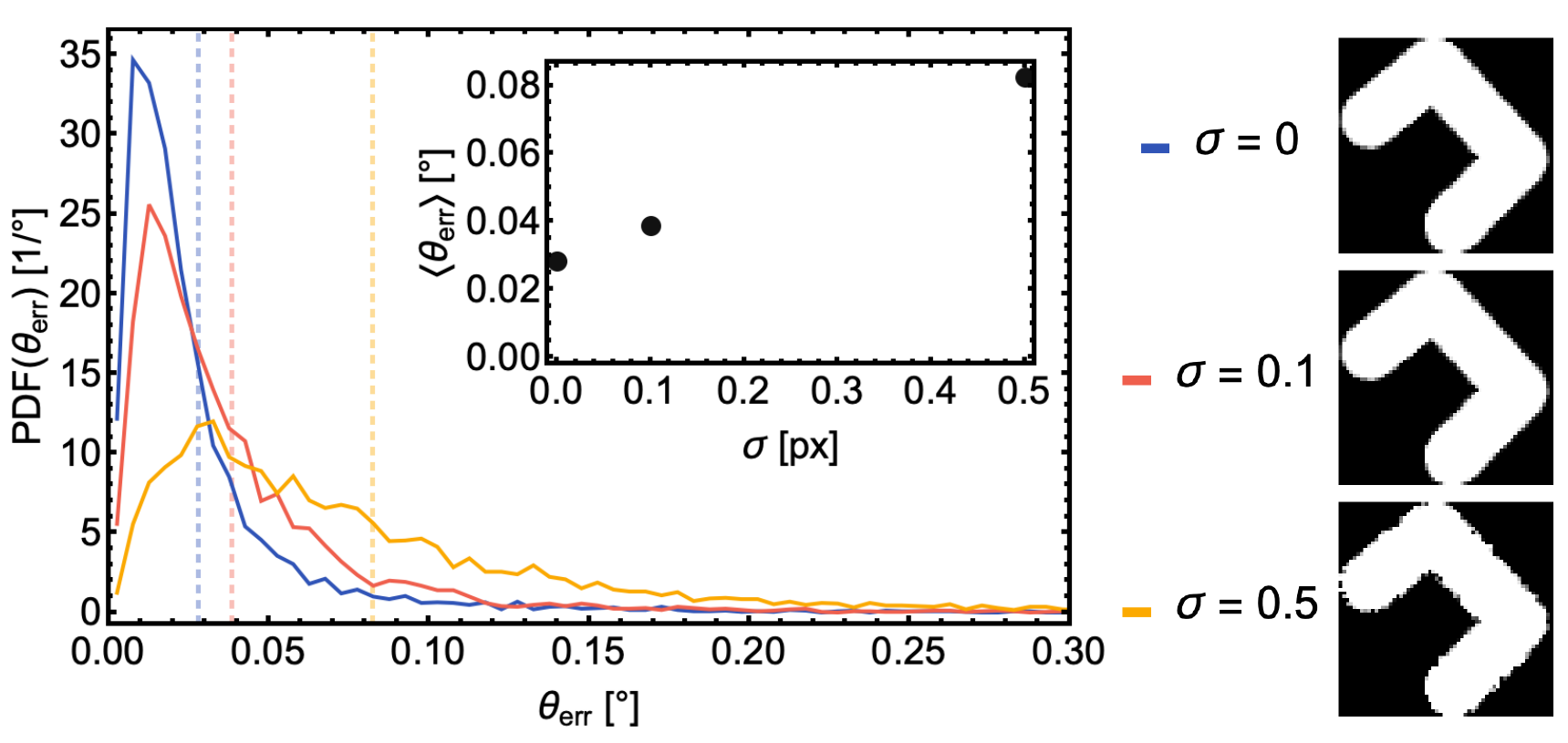}
    \caption{PDFs of the orientation error for a series of 5000 synthetic images of 60 px $\times$ 60 px, using 4 cameras arranged as in Figure \ref{fig:particle_projections}. The edge of the particle has Gaussian noise added with $\sigma$ variance as shown in the images on the right-hand side. The dashed lines show the mean orientation errors.}
    \label{fig:Noise_influence}
\end{figure*}

Finding the effect of noise is done by generating synthetic data and adding Gaussian noise. In this case we investigate the effect of Gaussian noise at the edge of the particle, simulating a slightly out of focus or poorly binarised particle as seen in Figure \ref{fig:Noise_influence}. The edge noise of the particles is added by isolating the greyscale pixels (not fully black or white) in the synthetic image and adding Gaussian noise to this part of the image. The intensity of the noise is varied by changing the variance of the Gaussian noise, denoted by $\sigma$. A set of 5000 random orientations is used to generate synthetic particle images, where the images are sized 60 px $\times$ 60 px. The camera orientation is set up as in Figure \ref{fig:particle_projections}, where all four cameras are used. The particles affected by noise and the corresponding PDFs of the orientation errors are shown in Figure \ref{fig:Noise_influence}. Here the vertical dashed lines show the mean orientation error for each noise level.
The PDFs of the orientation errors demonstrate that the orientation detection is very accurate overall, with small errors in the detected orientation. Here we note that the orientation error depends on how 'jagged' the particle silhouette is; more edges and corners lead to a smaller orientation error, whereas a more 'smooth' particle silhouette gives a slightly larger error. The edge noise has a clear effect on the orientation errors, but the method is robust enough to still be very accurate in finding the orientation. The mean orientation errors for this synthetic dataset using our tracking method are of order $\mathcal{O}(0.1^\circ)$ or less, which is comparable to or smaller than the errors in similar orientation tracking methods \cite{Zimmermann2011,Mathai2016}. The peaks in the orientation error PDFs are not located at $0^\circ$, which can be explained by the limited resolution of the images. As a result, two very close (but distinct) orientations give the same projections at low resolutions. Therefore, there exists an upper limit for the accuracy of this method, beyond which the reference image and the optimised image cannot be distinguished. The value of this upper limit is dependent on the image resolution, where a higher resolution image leads to lower errors at the accuracy limit, as illustrated in the section below.

\subsection{Image size}
\begin{figure}
    \centering
    \includegraphics[width = 0.45\textwidth]{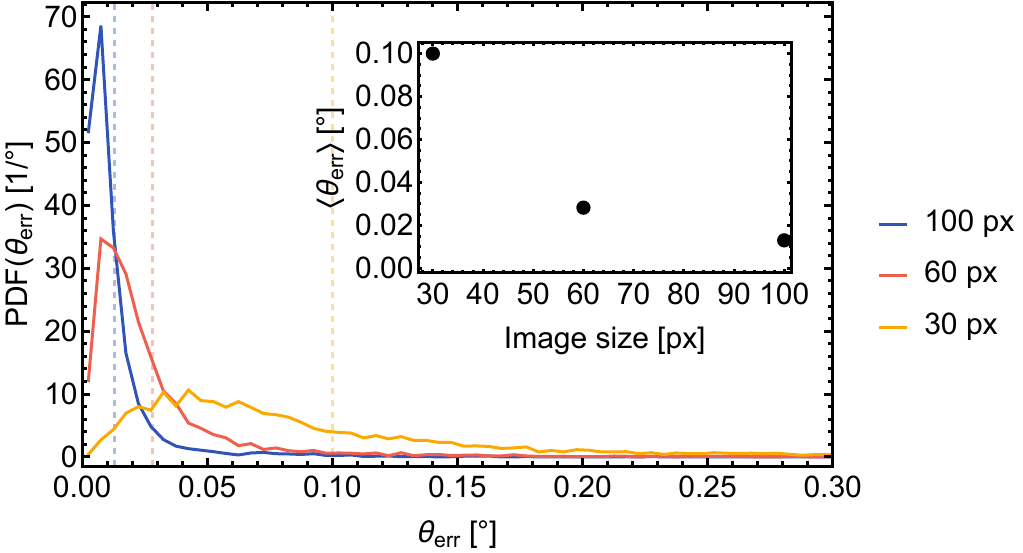}
    \caption{PDFs of the orientation error for different synthetic image sizes. 5000 random orientations are used for each size, where no noise is added. Four cameras are used, in the arrangement shown in Figure \ref{fig:particle_projections}. Dashed lines denote the mean orientation error.}
    \label{fig:Imagesize_influence}
\end{figure}

Similar to the effects of noise, we find the effect of the image size (now using no noise) using synthetic data. Synthetic particle images with particles of size \unit{30}{px} $\times$ \unit{30}{px}, \unit{60}{px} $\times$ \unit{60}{px}, and \unit{100}{px} $\times$ \unit{100}{px} are generated for 5000 random orientations. All four cameras are used as before (in the orientation of Figure \ref{fig:particle_projections}), and the error is computed for 5000 random orientations for each image size. The resulting PDFs are shown in Figure \ref{fig:Imagesize_influence}, showing that a larger image shows a smaller orientation error than smaller images.
This is an unsurprising result, since two very close but distinct orientations might be imaged identically at lower resolutions, higher resolutions allow us to distinguish these close orientations. Nonetheless, using smaller images still leads to very accurate tracked orientations overall.

\subsection{Number of cameras}
\begin{figure}
    \centering
    \includegraphics[width = 0.45\textwidth]{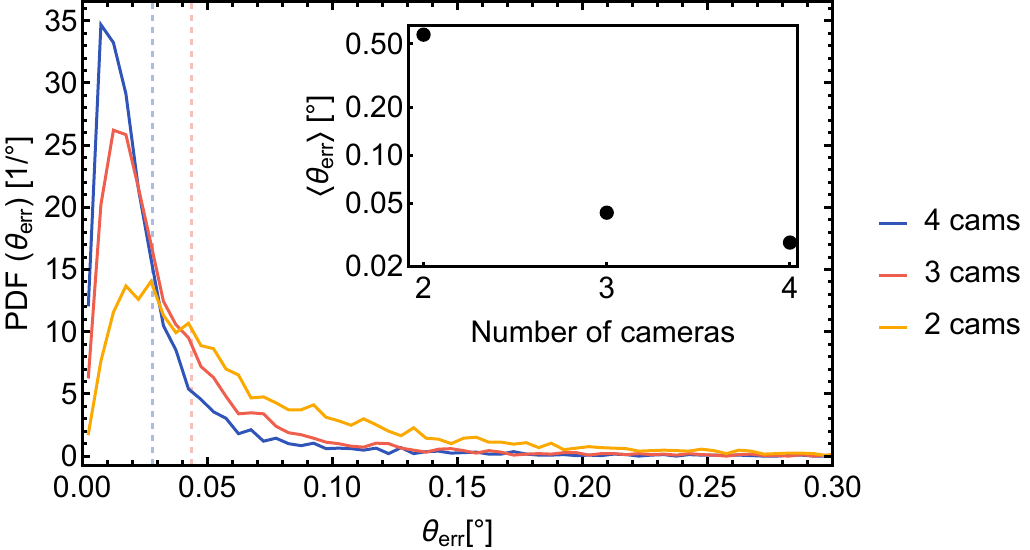}
    \caption{PDFs of the orientation error for 5000 random orientations using a different number of cameras. The images are sized 60 px $\times$ 60 px, with the mean orientation error marked by the dashed lines. No noise was added, and the camera setup as in Figure \ref{fig:particle_projections} is used. Note that the vertical axis in the inset is logarithmic.}
    \label{fig:Numcams_influence}
\end{figure}

So far, the orientation of particles is determined using 4 cameras, which we have claimed is done to improve accuracy. This section quantifies how well the method performs when varying the number of cameras used. The cameras are set up as shown in Figure \ref{fig:particle_projections}. For the 3 camera case, we use cameras 1 through 3, for 2 cameras, we use cameras 1 and 2. The effect of using a different number of cameras on the orientation errors is shown in Figure \ref{fig:Numcams_influence}.
This shows that the method performs significantly better when using more cameras. Nonetheless, even for only two cameras (which is not enough to determine the particle chirality for specific particles and orientations), the orientation is overwhelmingly found with a small error. Here we should note that part of the data for two cameras (0.4\% of the 5000 orientations) gives an erroneous first guess, leading to large values of $\theta_{err}$. This part of the data significantly alters the mean value for $\theta_{err}$, explaining the high $\langle \theta_{err} \rangle$ value for 2 cameras. Besides simply changing the number of cameras used, the setup of the cameras can also be varied, as discussed in the section below.

\subsection{Camera arrangements} \label{subsec: Camera arrangements}
\begin{figure*}
    \centering
    \includegraphics[width=\textwidth]{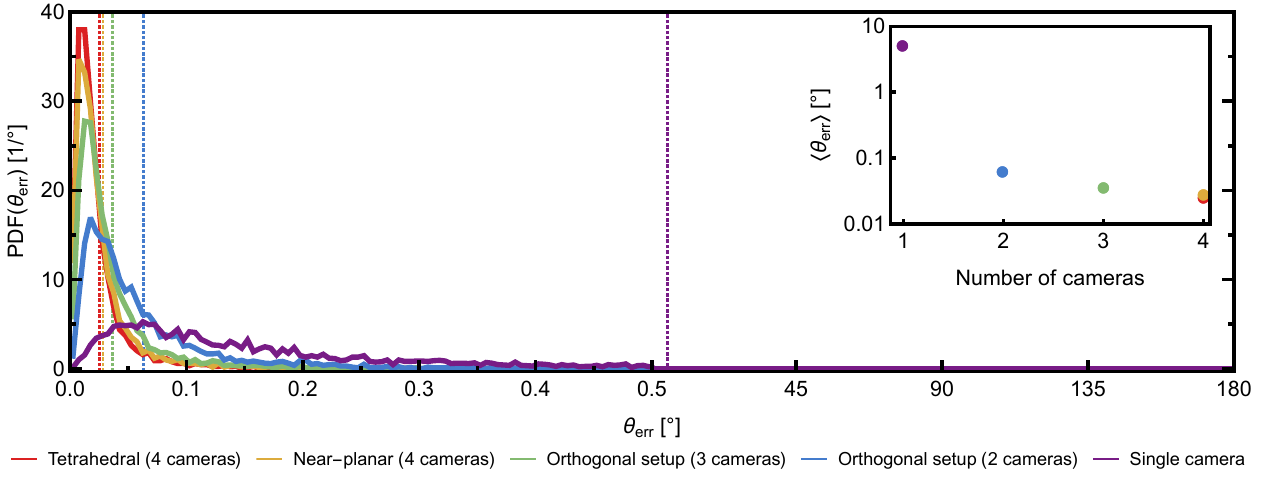}
    \caption{PDFs of the orientation error for chiral particle detection, using multiple camera arrangements. 5000 random orientations are used for each camera arrangement, where no noise is added, and the particle projections are sized \unit{60}{px} $\times$ \unit{60}{px}. The dashed lines show the weighted means of the orientation errors. Note that the horizontal axis uses two different linear scales. The inset shows the mean orientation error, where the vertical axis is logarithmic.}
    \label{fig:CamsOrient}
\end{figure*}

The camera setup used so far is the one used for the Taylor--Couette setup as seen in Figure \ref{fig:particle_projections}. These cameras are set up in an approximately planar arrangement (which we henceforth refer to as near-planar), imaging largely the same side of the particle. To quantitatively determine how much the camera setup affects the orientation error, we generate synthetic datasets of 5000 random orientations for different camera setups.
\begin{figure}
    \centering
    \includegraphics[width=0.25\textwidth]{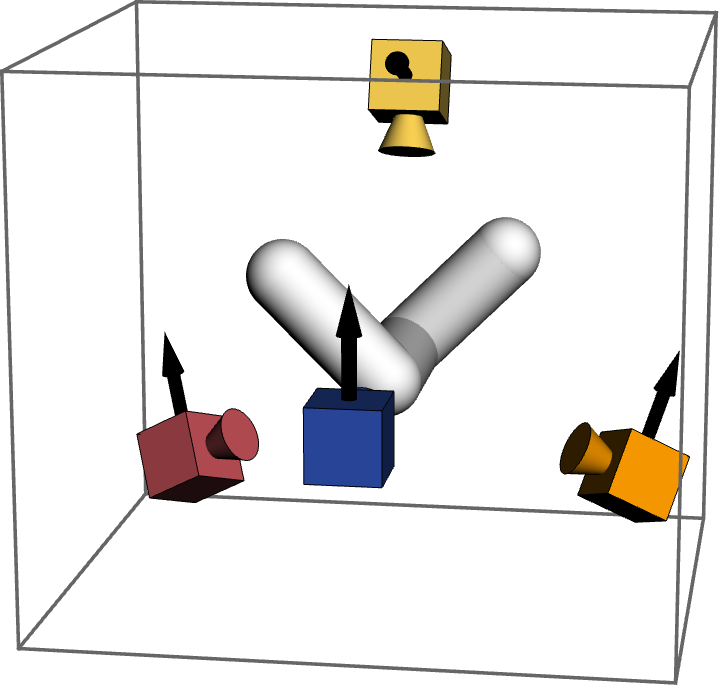}
    \caption{Tetrahedral camera setup using 4 cameras, imaging a chiral particle (figure not to scale).}
    \label{fig:Tetrahedral_setup}
\end{figure}
The investigated camera arrangements are near-planar, 2 cameras orthogonal, 3 orthogonal cameras, and a 4-camera tetrahedral arrangement as shown in Figure \ref{fig:Tetrahedral_setup} is tested. Finally, we test the accuracy of this method using a single camera. Since a single camera is unable to deduce the chirality, we provide the chirality to the algorithm in this case. The PDFs of the orientation errors are shown in Figure \ref{fig:CamsOrient}. 
We find that using more cameras is in general more accurate, noting that the way the cameras are arranged is also important. The PDFs show that the tetrahedral setup of cameras is more accurate than the near-planar camera setup. Therefore, we find that the way the cameras are arranged also affects the orientation error, but changing the number of used cameras has a greater effect. This is evident when comparing the near-planar arrangement (which uses 4 cameras) to the three orthogonal cameras, where the near-planar arrangement has a smaller average orientation error compared to the three orthogonal cameras.\\
Using a single camera, we notice that the value for $\langle \theta_{err} \rangle$ is larger than for other arrangements. This is caused by the algorithm finding a number of incorrect orientations, since the particle orientation cannot be determined for specific orientations, resulting in large values of $\theta_{err}$). Still, more than $90\%$ of the orientations are found with a small error ($\theta_{err} < 1$). This shows that the orientation finding method still works in principle, but cannot be reliably applied to this particle for a single camera.

\section{Examples of experimental applications}
\label{sec:Application}
\subsection{Chiral particle tracking}
\begin{figure}
    \centering
    \includegraphics[width = 0.45 \textwidth]{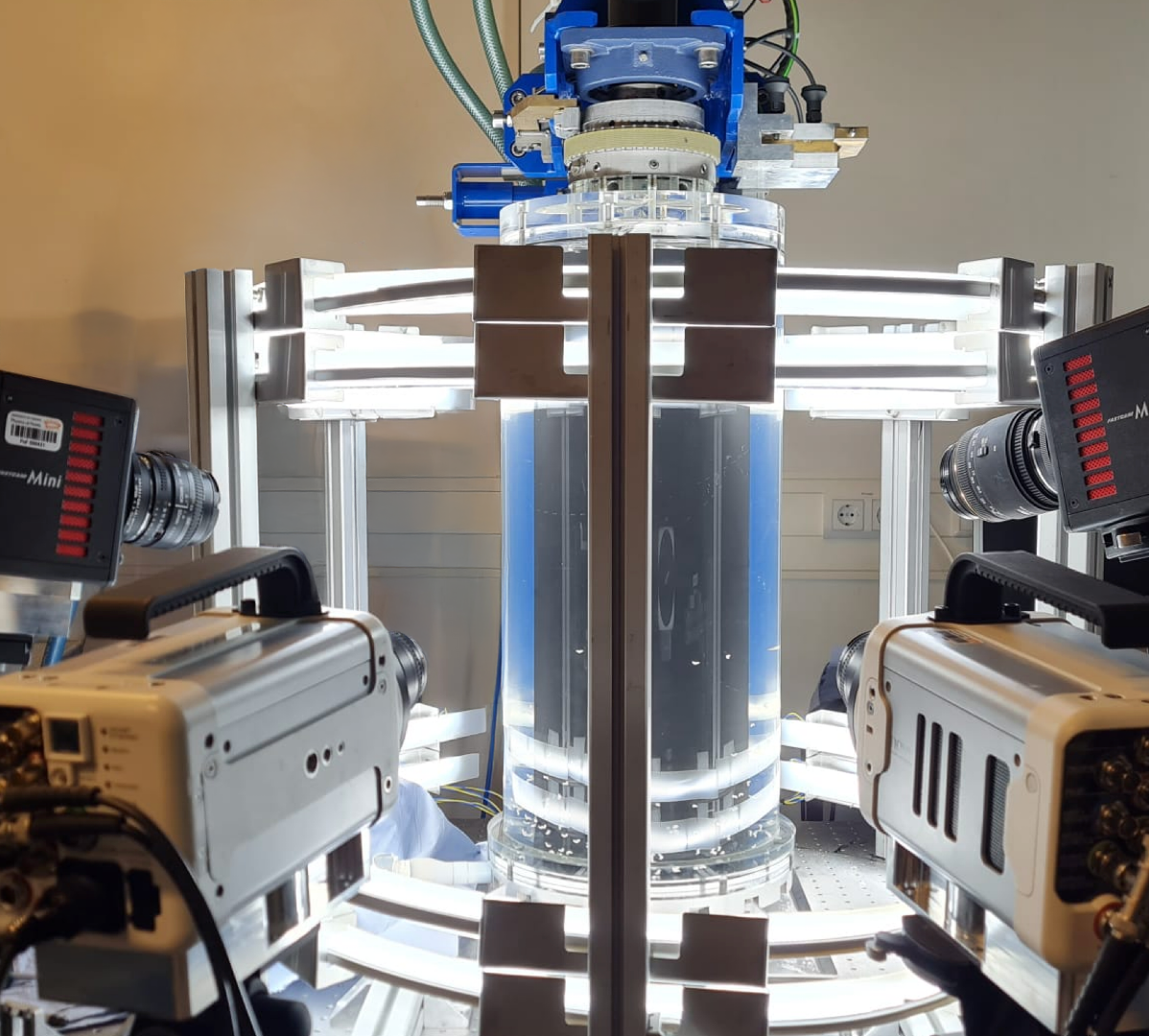}
    \caption{Photo of the experimental setup, where the four high-speed cameras image a small volume in the Taylor--Couette setup. The setup is characterised by the radial dimensions $r_i =$ \unit{75}{mm}, and $r_o =$ \unit{105}{mm}. The gap width is $d = r_o - r_i =$ \unit{30}{mm}, and the gap height is $L =$ \unit{549}{mm}, giving a volume $V =$ \unit{9.3}{L}.}
    \label{fig:Setup_camera}
\end{figure}

\begin{figure}
    \centering
    \includegraphics[width = 0.45 \textwidth]{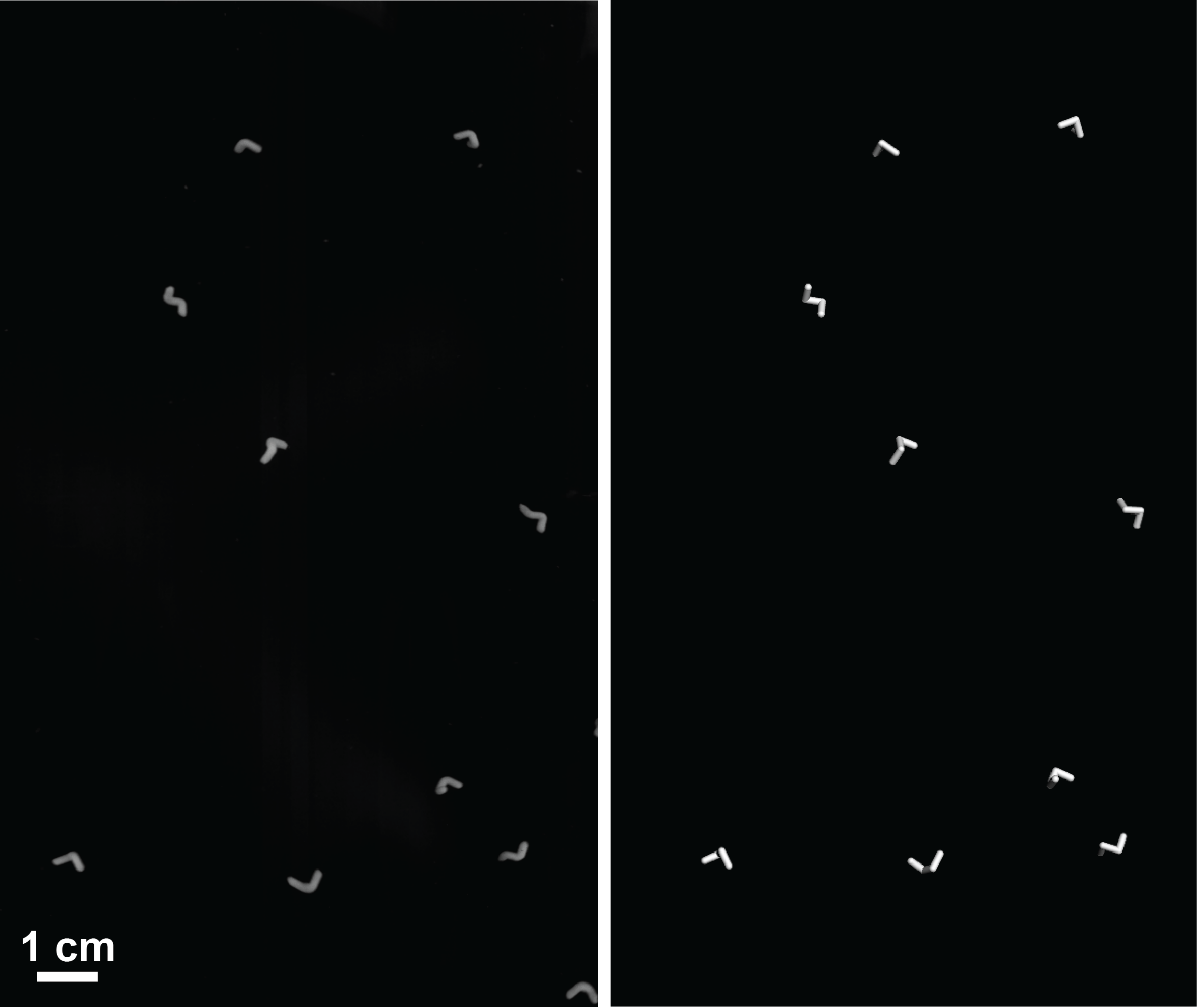}
    \caption{A cropped raw image captured by one of the cameras (left panel). A reconstruction of the particles in 3D (right panel).}
    \label{fig:Measurement_reconstruction}
\end{figure}

\begin{figure}
    \centering
    \includegraphics[width=0.45\textwidth]{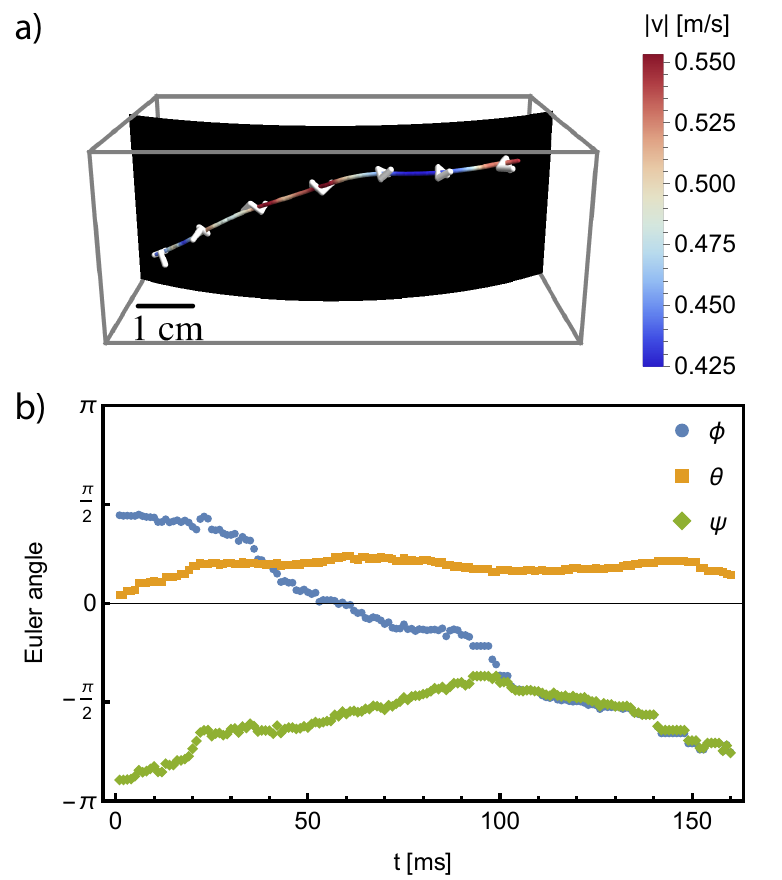}
    \caption{a) A single chiral particle shown at numerous time steps. The snapshots are spaced every \unit{0.025}{s}, and the measurement was performed at \unit{1000}{fps}. The grey curved surface indicates the inner cylinder of the Taylor--Couette. The coloured tube shows the centre of mass of the particle over time, where the colour denotes its speed. b) Computed Euler angles over time for a single particle trajectory. The order of rotations for the Euler angles is Z-Y-X.}
    \label{fig:Single_particle_track}
\end{figure}

The tracking method was used to track particles in a Taylor--Couette setup \cite{vanGils2011, Huisman2015} and the Dodecahedron setup, used for quiescent settling. The chiral particles and tetrads shown in Figure \ref{fig:Particles_photo} were tracked in the Taylor--Couette and Dodecahedron setup respectively. In the first experiment, a large number of chiral particles (of both chiralities) was added in the Boiling Twente Taylor-Couette setup \cite{Huisman2015} shown in Figure \ref{fig:Setup_camera}. The chiral particles are 3D printed by Shapeways using a sintering method, resulting in rough, slightly porous particles. These particles have a typical size of \unit{5}{mm}.
The setup has an inner cylinder radius $r_i =$ \unit{75}{mm}, and an outer cylinder radius $r_o =$ \unit{105}{mm}, resulting in a gap width $w = $ \unit{30}{mm}. The gap height is $L =$ \unit{549}{mm}, giving a volume $V =$ \unit{9.3}{L}.
To track the particles, we use 4 cameras: 2 Photron Nova S12 cameras (1024 px $\times$ 1024 px) and 2 Photron Mini AX-200 cameras (1024 px $\times$ 1024 px), each equipped with a \unit{50}{mm} lens, recording at \unit{1000}{fps}, with a resolution of approximately \unit{150}{\micro m \per px}. The imaged measurement volume (visible on all cameras) measures approximately \unit{10}{cm} $\times$ \unit{3}{cm} $\times$ \unit{10}{cm}. The cameras are set up in an almost planar manner around the Taylor--Couette setup as seen in Figure \ref{fig:Setup_camera}. The calibration is performed using ray matching \cite{Bourgoin2020}, after which the experiment is performed. The Taylor--Couette cylinders are rotate at \unit{5}{Hz} for the outer cylinder and \unit{-2}{Hz} for the inner cylinder, creating strong Taylor vortices \cite{vanderVeen2016}. The recorded images are processed using the algorithm described above, giving the location and orientation of the imaged chiral particles. A typical raw image and reconstruction using the found orientations is shown in Figure \ref{fig:Measurement_reconstruction}. The left panel shows the raw image of one of the cameras, whereas the right panel shows the reconstructed particles.
When comparing the raw image to the reconstruction, we see that not all particles from the raw image are reconstructed. This is due to some particles not being viewed by all four cameras, which we do not reconstruct. Additionally, overlapping particles are not reconstructed either, since we cannot disentangle the shapes in a reliable manner to find the orientation of individual particles in these cases.\\
The processed data gives the particle locations and orientation over time, allowing an investigation of the particle's rotation dynamics, which is the objective of many studies \cite{LaPorta2001,Calzavarini2009,Klein2013,Parsa2012}. Figure \ref{fig:Single_particle_track} shows the orientation and location of a single particle over time.
The light blue line shows the position of the particle's centre of mass over time, while the curved black backdrop represents the inner cylinder of the Taylor--Couette setup shown in Figure \ref{fig:Setup_camera}. The Euler angles over time are not smoothed, filtered, or post-processed, and show a few small jumps in the particle orientation (mainly in the $\phi$ component). These artefacts are caused by the optimisation loop finding a slightly inaccurate local minimum from the previous frames, then jumping to a better local minimum. This data could be processed further by smoothing the data, and optimising the orientation around the jumps again. Overall, the Euler angles change smoothly over time despite the turbulent flow the chiral particles are subjected to.

\subsection{Tetrads}
\begin{figure}
    \centering
    \includegraphics[width=0.45\textwidth]{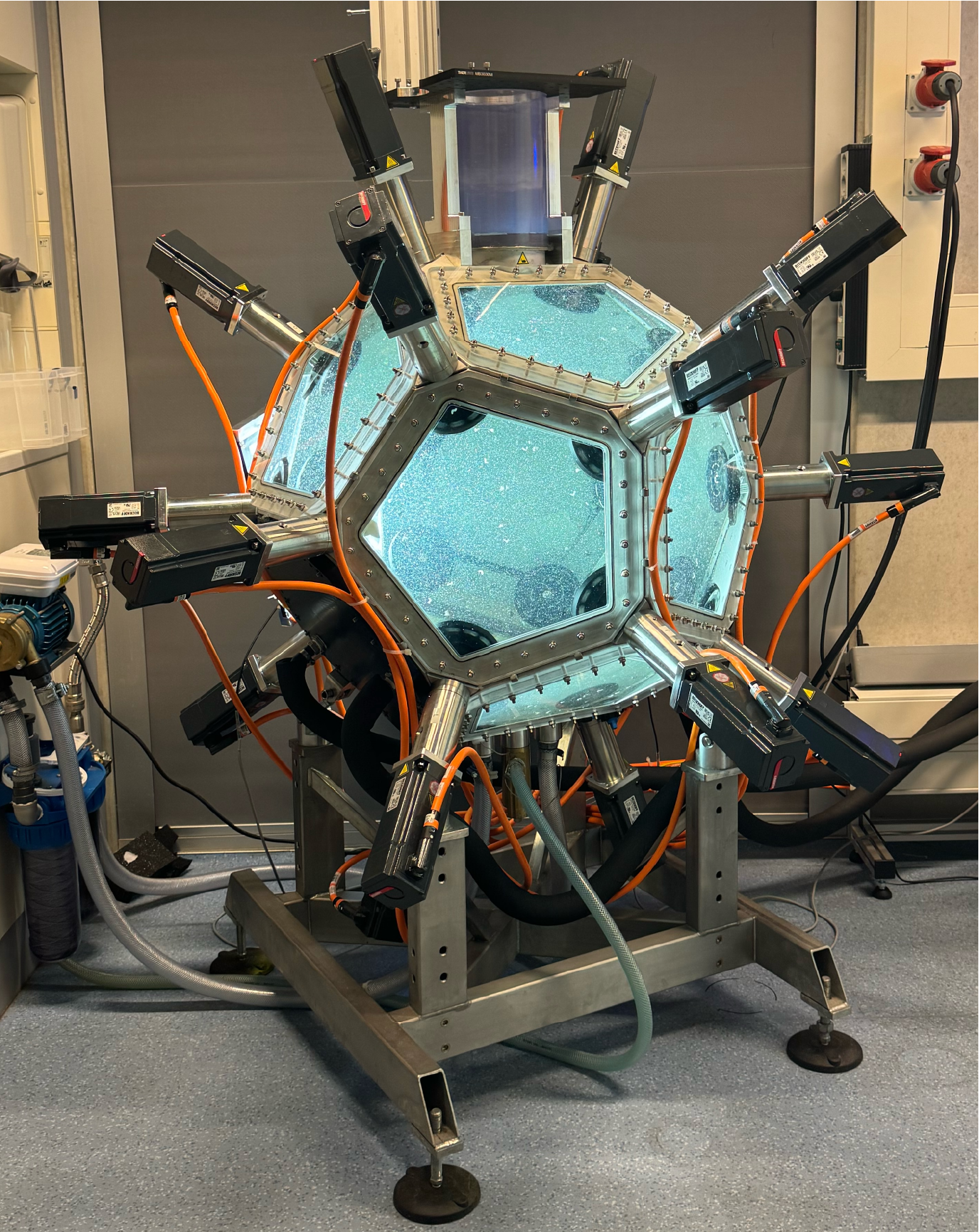}
    \caption{Dodecahedron setup used to track settling tetrads.}
    \label{fig:Dodecahedron setup}
\end{figure}

\begin{figure}
    \centering
    \includegraphics[width=\columnwidth]{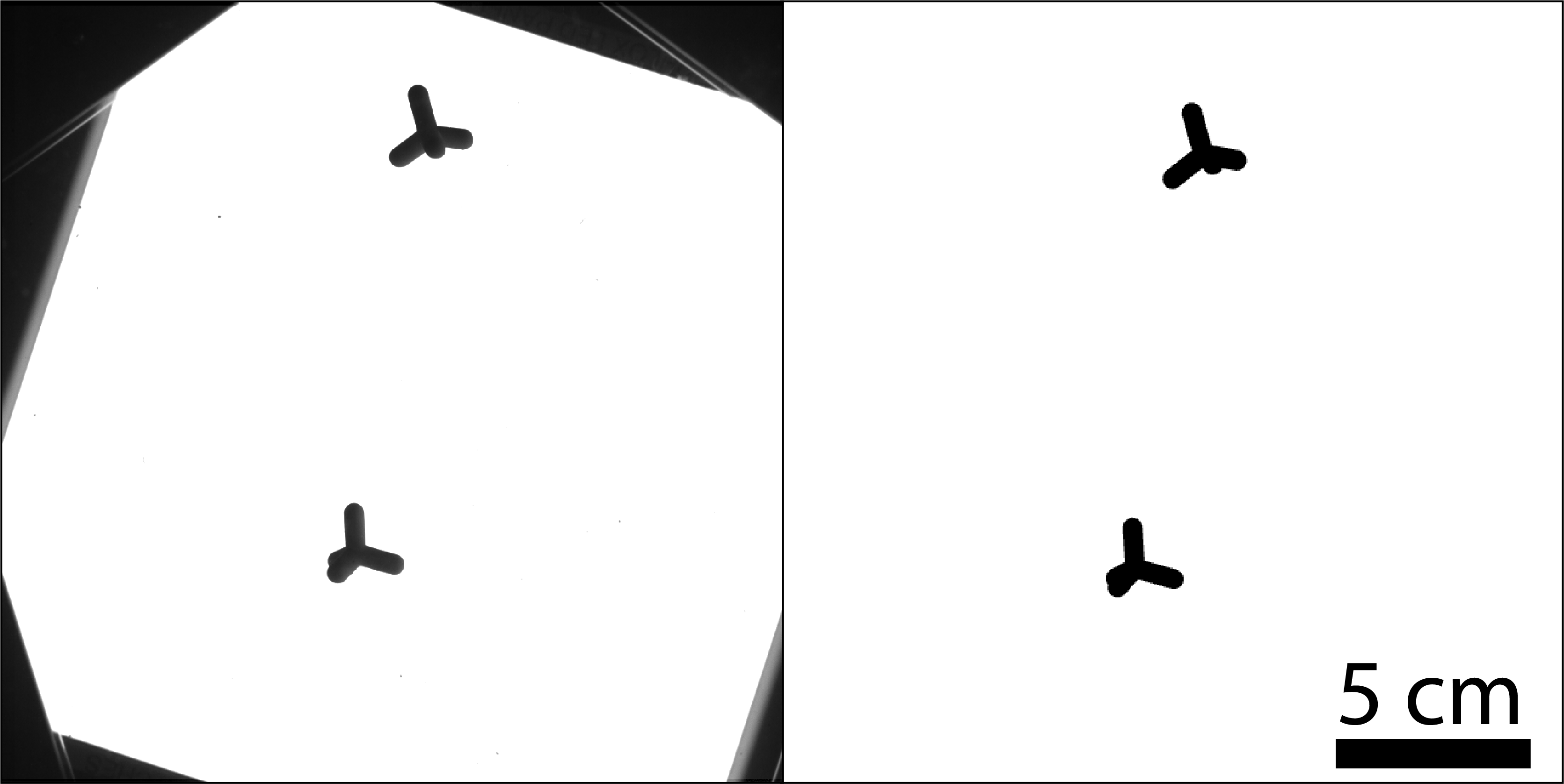}
    \caption{Raw image of settling tetrads in quiescent fluid in the left panel, recorded at 250 fps. The right panel shows the synthetically reconstructed tetrad particles.}
    \label{fig:Tetrad_reconstruction}
\end{figure}

\begin{figure}
    \centering
    \includegraphics[width=0.45\textwidth]{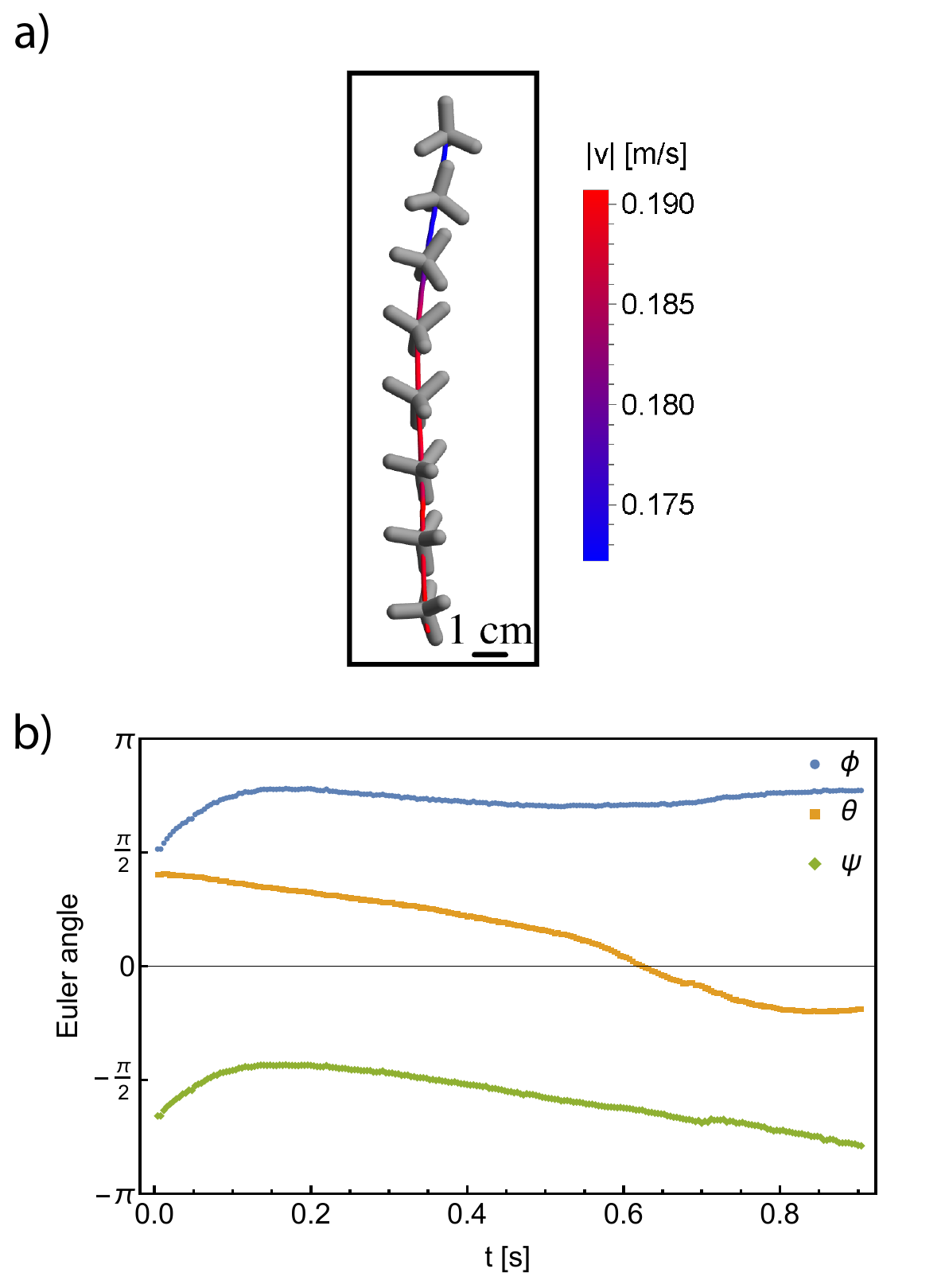}
    \caption{a) Snapshots of a single settling tetrad over time. The displayed snapshots are spaced \unit{0.12}{s} apart. The particle was recorded at 250 fps. The tube indicates the centre of mass of the particle, where the colour indicates its speed. b) The evolution of the Euler angles of the tracked tetrad particle over time. The order of rotations for the Euler angles is Z-Y-X.}
    \label{fig:Tetrad_track_angle}
\end{figure}

In the second application of the tracking method, we tracked settling tetrads in a Dodecahedron setup shown in Figure \ref{fig:Dodecahedron setup} (comparable to the Lagrangian exploration module \cite{Zimmermann2010}). The tetrads are 3D printed using Formlabs model resin, giving smooth particles with a typical length of \unit{2}{cm}.
This dodecahedral setup has edge lengths of \unit{40}{cm}, giving a volume of approximately \unit{210}{L}. The particles are tracked using 3 Photron Mini AX-200 cameras (1024 px $\times$ 1024 px), which image a measurement volume of \unit{10}{cm} $\times$ \unit{20}{cm} $\times$ \unit{20}{cm}. The camera resolution is approximately \unit{200}{\micro m \per px}. The imaged particles are generally around \unit{120}{px} $\times$ \unit{120}{px} in size. The experiments are recorded at \unit{250}{fps}, where the particles settle one by one in a quiescent fluid. In contrast to the previous application, the particles are backlit, opposed to the frontlit particles in the Taylor--Couette setup.\\
Similar to before, we use our algorithm to find the orientation of the tetrads. A raw image is shown alongside a reconstructed particle in Figure \ref{fig:Tetrad_reconstruction}. This demonstrates that the particle orientation can be found for different particle shapes, also for particles with a higher number of rotational symmetries.
The reconstructed particles again show a good visual match to the experimental images. To check the quality and accuracy of the tracked orientations, we find the orientation over time of a single particle. Figure \ref{fig:Tetrad_track_angle}a shows snapshots of a reconstructed settling tetrad over time. The cyan line indicates the centre of mass. Figure \ref{fig:Tetrad_track_angle}b shows the Euler angles of this settling tetrad over time.
The Euler angle curves are smooth over time, with only a slight jump in the angles around \unit{0.7}{s}. This jump is caused by the particle going out of frame on one of the cameras. This figure demonstrates that the particle orientation can be tracked smoothly if the image resolution and time resolution are sufficiently high.

\subsection{Oloids}
\begin{figure}
    \centering
    \includegraphics[width=\columnwidth]{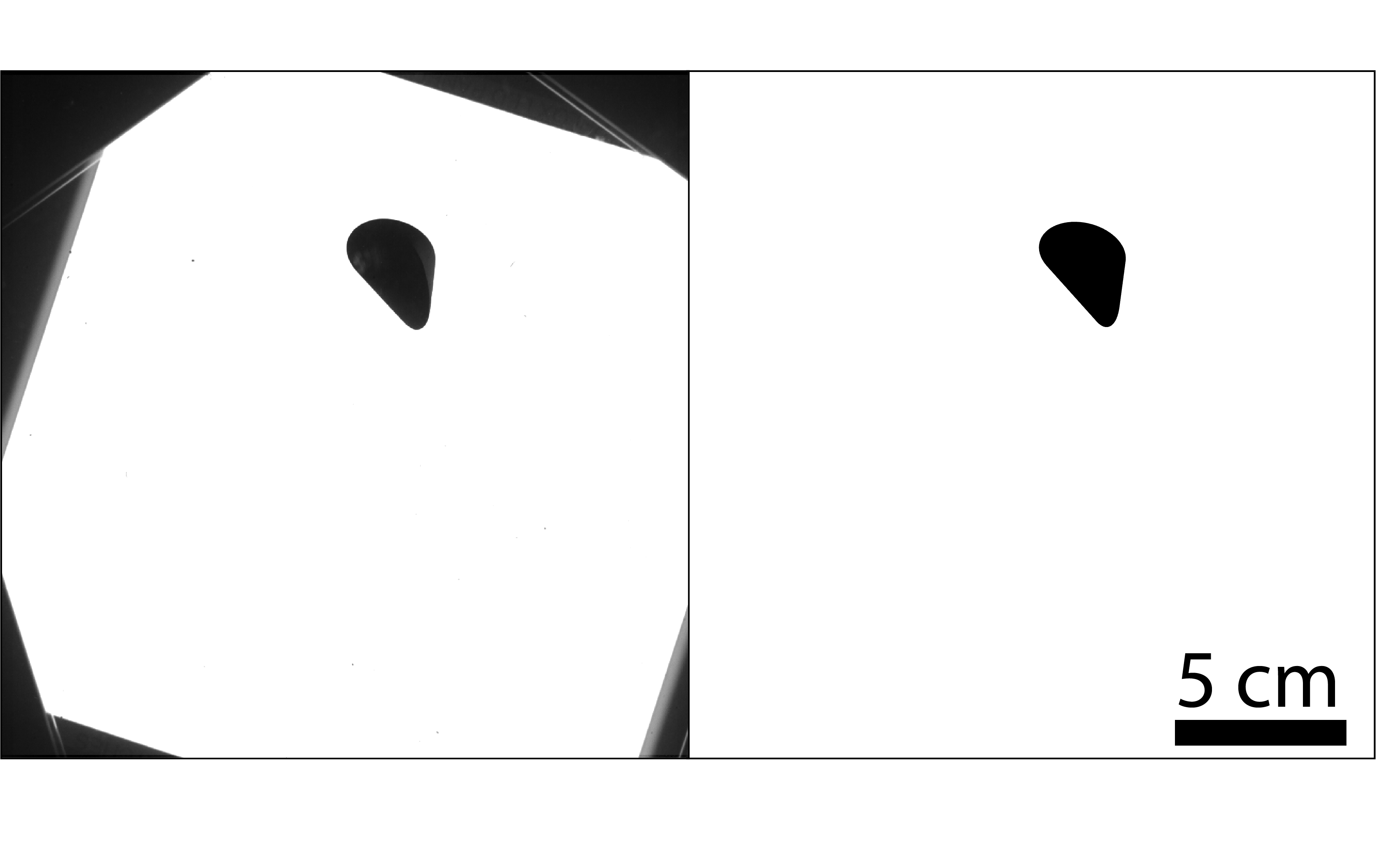}
    \caption{Raw image of a settling oloid in the left panel, recorded at 250 fps. Reconstructed oloid using the orientation tracking method shown in the right panel.}
    \label{fig:Oloid_reconstruction}
\end{figure}

\begin{figure}
    \centering
    \includegraphics[width=0.45\textwidth]{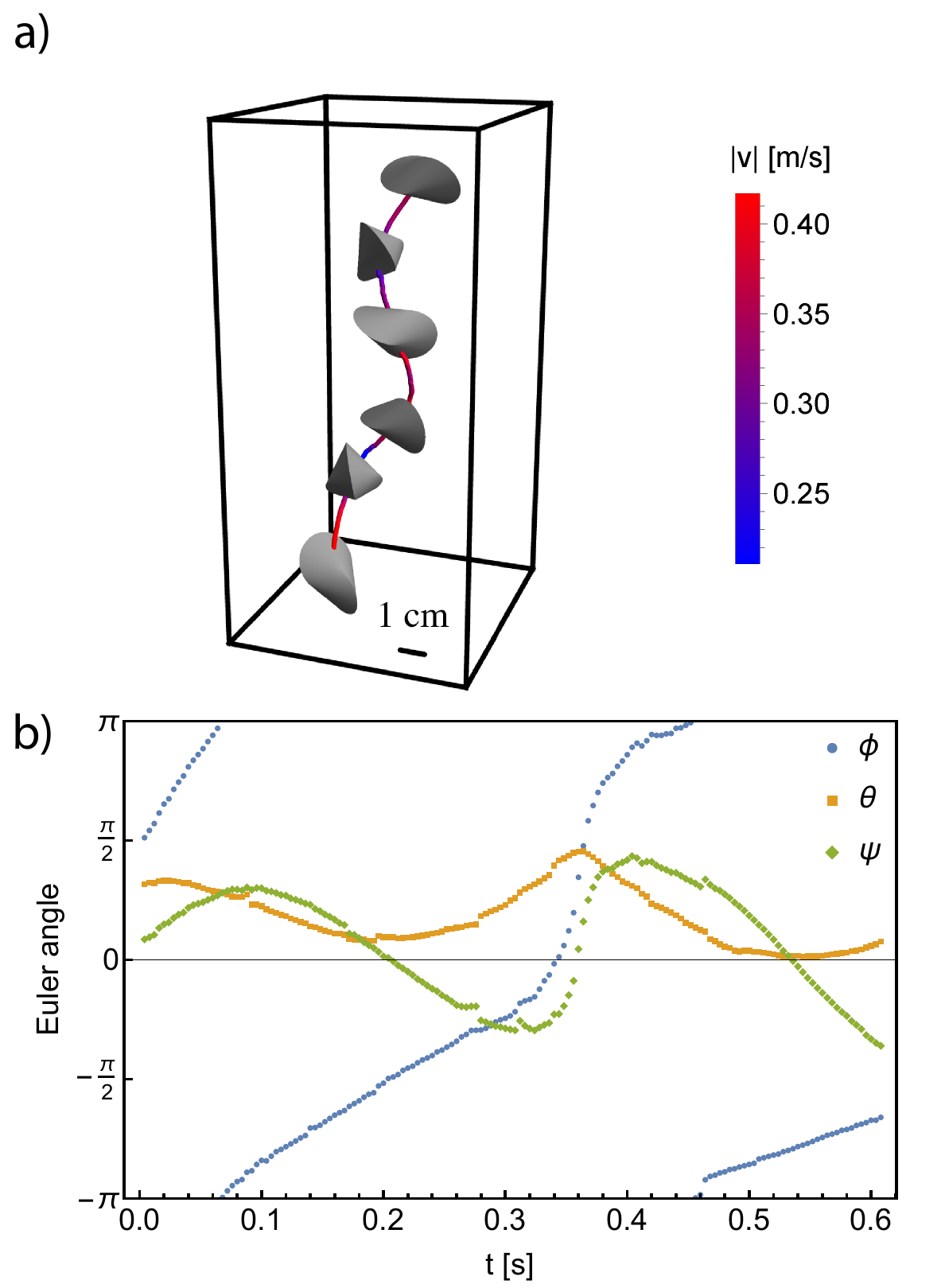}
    \caption{a) Snapshots of a settling oloid over time, spaced \unit{0.12}{s} apart (recorded at \unit{250}{fps}). The coloured tube shows the position of the centre of mass, where the colour indicates its speed. b) Plot of the Euler angles of the oloid's reconstructed orientation over time. The order of rotations for the Euler angles is Z-Y-X.}
    \label{fig:Oloid_track_angle}
\end{figure}

An oloid is a much different shape compared to the previously investigated tetrad and chiral particle, since the oloid is a convex shape, as opposed to the convave tetrads and chiral particles. Also, the particles are different in the sense that the oloid is not constructed from multiple tubes, which the chiral particles and tetrads are. Settling oloids are tracked in a quiescent fluid as was done for the tetrad particles. The orientation of the oloid is found using the orientation tracking method described above. To generate a synthetic particle projection, a slight change was made in the algorithm. Instead of generating the synthetic particle projection by drawing lines from vertex to vertex, we generate the two circles of the oloid using a number of points. These points on the circles are rotated and projected for each specific camera. Taking the convex hull of the projected points then gives the projection of the oloid for each camera. The orientation tracking method can then be used as before to determine the orientation of the settling oloid.\\
A recorded image and its reconstruction are shown in Figure \ref{fig:Oloid_reconstruction}, showing good agreement between the experimental image and the reconstructed particle.
The used orientation tracking algorithm therefore works well for a wide range of different particles, as long as the particle orientation can be derived from its projection, and the particle projections can be accurately generated. The tracked oloid particle over time is shown in Figure \ref{fig:Oloid_track_angle}.
The tracked orientation over time shows a periodic motion of the oloid. Even though there are slight discontinuities in some Euler angle curves (for example because of the particle going out of frame for one of the cameras), the general orientation dynamics are clearly visible with this tracking method. This again affirms that the used orientation tracking algorithm can be used to study the rotation and orientation dynamics of widely varying particles.

\section{Outlook \& Conclusion}
\label{sec:Conclusion}
The described orientation tracking algorithm is shown to be versatile and can be used to track various anisotropic particles using multiple high-speed cameras. The cameras can be set up in a flexible manner: from several cameras perpendicular, a tetrahedral configuration, to a near-planar arrangement. Using synthetic data, the accuracy of the method is quantified, and its robustness is evaluated. The method is shown to be more accurate when using more cameras, and when imaging the particles in higher resolution. The particles' orientation can still be determined when the particles are noisy, albeit somewhat less accurately. The arrangement of cameras is shown to affect the accuracy of the orientation determination. We verify and showcase our method by tracking the location and orientation of multiple types of particles simultaneously in different setups.\\
Overall, the described tracking method can be used for a vast range of particles, so long as the particle is anisotropic enough to determine its orientation from the projected shape. Furthermore, this method is flexible in its setup, where the cameras don't require any perpendicular angles or specific alignment, making it viable for imaging in setups with limited optical access.

\section{Supplementary material}
The supplementary material includes animations of the tracked particles, and shows raw videos of the particles alongside the 3D reconstruction made using the algorithm described in this manuscript. Additionally, the supplementary material contains the orientation tracking code described above, along with raw images and sample data. This allows one to find the orientation of the chiral particles in the sample data. The full datasets are available upon reasonable request.

\begin{acknowledgments}
The authors thank Detlef Lohse and Federico Toschi for their insights and discussions. We would like to thank Gert-Wim Bruggert, Thomas Zijlstra, and Martin Bos for their technical support. This research has been funded by the Dutch Research Council (NWO) under grant OCENW.GROOT.2019.031. Elian Bernard visited the Physics of Fluids department under an Erasmus grant. The authors report no conflict of interest.
\end{acknowledgments}

\appendix
\section{Error propagation between position and orientation}
\label{sec:Error propagation}
\begin{figure}
    \centering
    \includegraphics[width=0.9\linewidth]{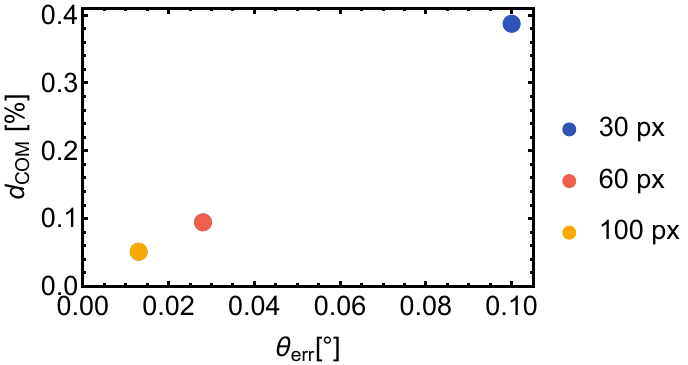}
    \caption{The error in the projected centre of mass as a percentage of the image size, as a function of the orientation error. Each datapoint represents 1000 orientations, each randomly rotated by the orientation error 100 times. The camera setup as in Figure \ref{fig:particle_projections} is used. The mean orientation error for the different image sizes as in Figure \ref{fig:Imagesize_influence} is used.}
    \label{fig:pos_error_from_orientation}
\end{figure}

The described algorithm treats the finding of the position and the finding of the orientation separately. However, the error in the position slightly affects the orientation, in the sense that the position determines the camera vector along which the synthetic particle model is projected.\\
For the experiment tracking many chiral particles, as shown in section \ref{sec:Application}, the average image size is approximately \unit{30}{px} $\times$ \unit{30}{px}. An error in the position of $20\%$ of the image size (which is approximately the biggest distance between the projected centre of mass, and the centroid of the projection) leads to a difference in camera vector of approximately $0.1^{\circ}$. This projection error leads to an orientation error of similar magnitude, which is comparable to the mean orientation error of the optimisation process. \\
This error is sufficiently small for the fluid-dynamics applications this method was developed for. If a higher accuracy is required, more iterations of the algorithm can be used, or the location and orientation can be optimised simultaneously.\\
\\
Since the orientation is used to correct the centre of mass position, the error in orientation affects the error in position. We test the effect of the orientation error by calculating the displacement of the projected centre of mass (in pixels), as a result of an erroneous orientation.\\
A set of 1000 random orientations is generated, where each orientation is randomly rotated 100 times by the orientation error. This calculation is done for three different square image sizes: \unit{30}{px}, \unit{60}{px}, and \unit{100}{px}, where we use the corresponding mean orientation error as found in Figure \ref{fig:Imagesize_influence} to rotate the reference orientations. The displacement of the projected centre of mass (in pixels) is then calculated for each camera. The displacements are averaged per image size, and calculated as a percentage of the image size: these results are shown in Figure \ref{fig:pos_error_from_orientation}.
This figure shows that the projected centre of mass is displaced by less than $0.5\%$ of the image size.\\
For the shown chiral particle setup, this gives a position error of $\mathcal{O}$(\unit{10}{\micro \meter}) resulting from the orientation error. The magnitude of this error is sufficiently small for the fluid-dynamics applications for which this algorithm was developed.

\section{Orientation errors for different particles}
\label{sec:Error particles}
\begin{figure}
    \centering
    \includegraphics[width=0.9\linewidth]{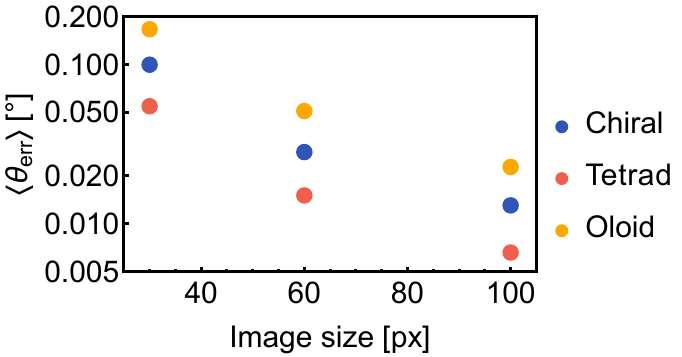}
    \caption{Mean orientation errors for different particle shapes at multiple image sizes. The camera setup as in Figure \ref{fig:particle_projections} is used, where no noise is added to the particles. The chiral particle data is the same as in Figure \ref{fig:Imagesize_influence}, based on 5000 random orientations for each image size. The tetrad and oloid shapes are both based on 1000 random orientations. Note that the vertical axis is logarithmically scaled.}
    \label{fig:Orient_error_geometries}
\end{figure}

So far, the robustness and the orientation errors have only been evaluated for the chiral particles. Here we evaluate the orientation errors for the other shapes in this manuscript (the tetrad and oloid), for different image sizes. The synthetic data is generated as before, now using 1000 orientations to determine the mean orientation error. The mean orientation errors for all used shapes are shown in Figure \ref{fig:Orient_error_geometries}, for multiple image sizes.
This plot shows that the magnitude of the mean orientation error differs slightly per particle shape, but the overall scaling with image size is similar for all shapes. The mean orientation errors are sufficiently small for our applications for all particle shapes.


\begin{thebibliography}{55}%
\makeatletter
\providecommand \@ifxundefined [1]{%
 \@ifx{#1\undefined}
}%
\providecommand \@ifnum [1]{%
 \ifnum #1\expandafter \@firstoftwo
 \else \expandafter \@secondoftwo
 \fi
}%
\providecommand \@ifx [1]{%
 \ifx #1\expandafter \@firstoftwo
 \else \expandafter \@secondoftwo
 \fi
}%
\providecommand \natexlab [1]{#1}%
\providecommand \enquote  [1]{``#1''}%
\providecommand \bibnamefont  [1]{#1}%
\providecommand \bibfnamefont [1]{#1}%
\providecommand \citenamefont [1]{#1}%
\providecommand \href@noop [0]{\@secondoftwo}%
\providecommand \href [0]{\begingroup \@sanitize@url \@href}%
\providecommand \@href[1]{\@@startlink{#1}\@@href}%
\providecommand \@@href[1]{\endgroup#1\@@endlink}%
\providecommand \@sanitize@url [0]{\catcode `\\12\catcode `\$12\catcode
  `\&12\catcode `\#12\catcode `\^12\catcode `\_12\catcode `\%12\relax}%
\providecommand \@@startlink[1]{}%
\providecommand \@@endlink[0]{}%
\providecommand \url  [0]{\begingroup\@sanitize@url \@url }%
\providecommand \@url [1]{\endgroup\@href {#1}{\urlprefix }}%
\providecommand \urlprefix  [0]{URL }%
\providecommand \Eprint [0]{\href }%
\providecommand \doibase [0]{http://dx.doi.org/}%
\providecommand \selectlanguage [0]{\@gobble}%
\providecommand \bibinfo  [0]{\@secondoftwo}%
\providecommand \bibfield  [0]{\@secondoftwo}%
\providecommand \translation [1]{[#1]}%
\providecommand \BibitemOpen [0]{}%
\providecommand \bibitemStop [0]{}%
\providecommand \bibitemNoStop [0]{.\EOS\space}%
\providecommand \EOS [0]{\spacefactor3000\relax}%
\providecommand \BibitemShut  [1]{\csname bibitem#1\endcsname}%
\let\auto@bib@innerbib\@empty
\bibitem [{\citenamefont {Corbetta}\ \emph {et~al.}(2018)\citenamefont
  {Corbetta}, \citenamefont {Meeusen}, \citenamefont {Lee}, \citenamefont
  {Benzi},\ and\ \citenamefont {Toschi}}]{Corbetta2018}%
  \BibitemOpen
  \bibfield  {author} {\bibinfo {author} {\bibfnamefont {A.}~\bibnamefont
  {Corbetta}}, \bibinfo {author} {\bibfnamefont {J.~A.}\ \bibnamefont
  {Meeusen}}, \bibinfo {author} {\bibfnamefont {C.}~\bibnamefont {Lee}},
  \bibinfo {author} {\bibfnamefont {R.}~\bibnamefont {Benzi}}, \ and\ \bibinfo
  {author} {\bibfnamefont {F.}~\bibnamefont {Toschi}},\ }\href@noop {}
  {\bibfield  {journal} {\bibinfo  {journal} {Phys. Rev. E}\ }\textbf {\bibinfo
  {volume} {98}} (\bibinfo {year} {2018})}\BibitemShut {NoStop}%
\bibitem [{\citenamefont {Garcimart\'{\i}n}\ \emph {et~al.}(2015)\citenamefont
  {Garcimart\'{\i}n}, \citenamefont {Pastor}, \citenamefont {Ferrer},
  \citenamefont {Ramos}, \citenamefont {Mart\'{\i}n-G\'omez},\ and\
  \citenamefont {Zuriguel}}]{Garcimartin2015}%
  \BibitemOpen
  \bibfield  {author} {\bibinfo {author} {\bibfnamefont {A.}~\bibnamefont
  {Garcimart\'{\i}n}}, \bibinfo {author} {\bibfnamefont {J.~M.}\ \bibnamefont
  {Pastor}}, \bibinfo {author} {\bibfnamefont {L.~M.}\ \bibnamefont {Ferrer}},
  \bibinfo {author} {\bibfnamefont {J.~J.}\ \bibnamefont {Ramos}}, \bibinfo
  {author} {\bibfnamefont {C.}~\bibnamefont {Mart\'{\i}n-G\'omez}}, \ and\
  \bibinfo {author} {\bibfnamefont {I.}~\bibnamefont {Zuriguel}},\ }\href@noop
  {} {\bibfield  {journal} {\bibinfo  {journal} {Phys. Rev. E}\ }\textbf
  {\bibinfo {volume} {91}} (\bibinfo {year} {2015})}\BibitemShut {NoStop}%
\bibitem [{\citenamefont {Salmon}\ \emph {et~al.}(2023)\citenamefont {Salmon},
  \citenamefont {Baker}, \citenamefont {Kozarek},\ and\ \citenamefont
  {Coletti}}]{Salmon2023}%
  \BibitemOpen
  \bibfield  {author} {\bibinfo {author} {\bibfnamefont {H.~R.~S.}\
  \bibnamefont {Salmon}}, \bibinfo {author} {\bibfnamefont {L.~J.}\
  \bibnamefont {Baker}}, \bibinfo {author} {\bibfnamefont {J.~L.}\ \bibnamefont
  {Kozarek}}, \ and\ \bibinfo {author} {\bibfnamefont {F.}~\bibnamefont
  {Coletti}},\ }\href@noop {} {\bibfield  {journal} {\bibinfo  {journal} {Water
  Resour. Res.}\ }\textbf {\bibinfo {volume} {59}} (\bibinfo {year}
  {2023})}\BibitemShut {NoStop}%
\bibitem [{\citenamefont {Schröder}\ and\ \citenamefont
  {Schanz}(2022)}]{Schroder2022}%
  \BibitemOpen
  \bibfield  {author} {\bibinfo {author} {\bibfnamefont {A.}~\bibnamefont
  {Schröder}}\ and\ \bibinfo {author} {\bibfnamefont {D.}~\bibnamefont
  {Schanz}},\ }\href@noop {} {\bibfield  {journal} {\bibinfo  {journal} {Annu.
  Rev. Fluid Mech.}\ }\textbf {\bibinfo {volume} {55}} (\bibinfo {year}
  {2022})}\BibitemShut {NoStop}%
\bibitem [{\citenamefont {Godbersen}\ \emph {et~al.}(2021)\citenamefont
  {Godbersen}, \citenamefont {Bosbach}, \citenamefont {Schanz},\ and\
  \citenamefont {Schröder}}]{Godbersen2021}%
  \BibitemOpen
  \bibfield  {author} {\bibinfo {author} {\bibfnamefont {P.}~\bibnamefont
  {Godbersen}}, \bibinfo {author} {\bibfnamefont {J.}~\bibnamefont {Bosbach}},
  \bibinfo {author} {\bibfnamefont {D.}~\bibnamefont {Schanz}}, \ and\ \bibinfo
  {author} {\bibfnamefont {A.}~\bibnamefont {Schröder}},\ }\href@noop {}
  {\bibfield  {journal} {\bibinfo  {journal} {Phys. Rev. Fluids}\ }\textbf
  {\bibinfo {volume} {6}} (\bibinfo {year} {2021})}\BibitemShut {NoStop}%
\bibitem [{\citenamefont {Petersen}, \citenamefont {Baker},\ and\ \citenamefont
  {Coletti}(2019)}]{Petersen2019}%
  \BibitemOpen
  \bibfield  {author} {\bibinfo {author} {\bibfnamefont {A.~J.}\ \bibnamefont
  {Petersen}}, \bibinfo {author} {\bibfnamefont {L.}~\bibnamefont {Baker}}, \
  and\ \bibinfo {author} {\bibfnamefont {F.}~\bibnamefont {Coletti}},\
  }\href@noop {} {\bibfield  {journal} {\bibinfo  {journal} {J. Fluid Mech}\
  }\textbf {\bibinfo {volume} {864}} (\bibinfo {year} {2019})}\BibitemShut
  {NoStop}%
\bibitem [{\citenamefont {Nemes}\ \emph {et~al.}(2017)\citenamefont {Nemes},
  \citenamefont {Dasari}, \citenamefont {Hong}, \citenamefont {Guala},\ and\
  \citenamefont {Coletti}}]{Nemes2017}%
  \BibitemOpen
  \bibfield  {author} {\bibinfo {author} {\bibfnamefont {A.}~\bibnamefont
  {Nemes}}, \bibinfo {author} {\bibfnamefont {T.}~\bibnamefont {Dasari}},
  \bibinfo {author} {\bibfnamefont {J.}~\bibnamefont {Hong}}, \bibinfo {author}
  {\bibfnamefont {M.}~\bibnamefont {Guala}}, \ and\ \bibinfo {author}
  {\bibfnamefont {F.}~\bibnamefont {Coletti}},\ }\href@noop {} {\bibfield
  {journal} {\bibinfo  {journal} {J. Fluid Mech.}\ }\textbf {\bibinfo {volume}
  {814}} (\bibinfo {year} {2017})}\BibitemShut {NoStop}%
\bibitem [{\citenamefont {Collins}\ \emph {et~al.}(2021)\citenamefont
  {Collins}, \citenamefont {Hamati}, \citenamefont {Candelier}, \citenamefont
  {Gustavsson}, \citenamefont {Mehlig},\ and\ \citenamefont
  {Voth}}]{Collins2021}%
  \BibitemOpen
  \bibfield  {author} {\bibinfo {author} {\bibfnamefont {D.}~\bibnamefont
  {Collins}}, \bibinfo {author} {\bibfnamefont {R.~J.}\ \bibnamefont {Hamati}},
  \bibinfo {author} {\bibfnamefont {F.}~\bibnamefont {Candelier}}, \bibinfo
  {author} {\bibfnamefont {K.}~\bibnamefont {Gustavsson}}, \bibinfo {author}
  {\bibfnamefont {B.}~\bibnamefont {Mehlig}}, \ and\ \bibinfo {author}
  {\bibfnamefont {G.~A.}\ \bibnamefont {Voth}},\ }\href@noop {} {\bibfield
  {journal} {\bibinfo  {journal} {Phys. Rev. Fluids}\ }\textbf {\bibinfo
  {volume} {6}} (\bibinfo {year} {2021})}\BibitemShut {NoStop}%
\bibitem [{\citenamefont {Brown}, \citenamefont {Warhaft},\ and\ \citenamefont
  {Voth}(2009)}]{Brown2009}%
  \BibitemOpen
  \bibfield  {author} {\bibinfo {author} {\bibfnamefont {R.~D.}\ \bibnamefont
  {Brown}}, \bibinfo {author} {\bibfnamefont {Z.}~\bibnamefont {Warhaft}}, \
  and\ \bibinfo {author} {\bibfnamefont {G.~A.}\ \bibnamefont {Voth}},\
  }\href@noop {} {\bibfield  {journal} {\bibinfo  {journal} {Phys. Rev. Lett.}\
  }\textbf {\bibinfo {volume} {103}} (\bibinfo {year} {2009})}\BibitemShut
  {NoStop}%
\bibitem [{\citenamefont {Chiarini}\ and\ \citenamefont
  {Rosti}(2024)}]{Chiarini2024}%
  \BibitemOpen
  \bibfield  {author} {\bibinfo {author} {\bibfnamefont {A.}~\bibnamefont
  {Chiarini}}\ and\ \bibinfo {author} {\bibfnamefont {M.}~\bibnamefont
  {Rosti}},\ }\href@noop {} {\bibfield  {journal} {\bibinfo  {journal} {J.
  Fluid Mech.}\ }\textbf {\bibinfo {volume} {988}} (\bibinfo {year}
  {2024})}\BibitemShut {NoStop}%
\bibitem [{\citenamefont {Voth}\ and\ \citenamefont
  {Soldati}(2017)}]{Voth2017}%
  \BibitemOpen
  \bibfield  {author} {\bibinfo {author} {\bibfnamefont {G.~A.}\ \bibnamefont
  {Voth}}\ and\ \bibinfo {author} {\bibfnamefont {A.}~\bibnamefont {Soldati}},\
  }\href@noop {} {\bibfield  {journal} {\bibinfo  {journal} {Annu. Rev. Fluid
  Mech.}\ }\textbf {\bibinfo {volume} {49}} (\bibinfo {year}
  {2017})}\BibitemShut {NoStop}%
\bibitem [{\citenamefont {Bakhuis}\ \emph {et~al.}(2019)\citenamefont
  {Bakhuis}, \citenamefont {Mathai}, \citenamefont {Verschoof}, \citenamefont
  {Ezeta}, \citenamefont {Lohse}, \citenamefont {Huisman},\ and\ \citenamefont
  {Sun}}]{Bakhuis2019}%
  \BibitemOpen
  \bibfield  {author} {\bibinfo {author} {\bibfnamefont {D.}~\bibnamefont
  {Bakhuis}}, \bibinfo {author} {\bibfnamefont {V.}~\bibnamefont {Mathai}},
  \bibinfo {author} {\bibfnamefont {R.~A.}\ \bibnamefont {Verschoof}}, \bibinfo
  {author} {\bibfnamefont {R.}~\bibnamefont {Ezeta}}, \bibinfo {author}
  {\bibfnamefont {D.}~\bibnamefont {Lohse}}, \bibinfo {author} {\bibfnamefont
  {S.~G.}\ \bibnamefont {Huisman}}, \ and\ \bibinfo {author} {\bibfnamefont
  {C.}~\bibnamefont {Sun}},\ }\href@noop {} {\bibfield  {journal} {\bibinfo
  {journal} {Phys. Rev. Fluids}\ }\textbf {\bibinfo {volume} {4}} (\bibinfo
  {year} {2019})}\BibitemShut {NoStop}%
\bibitem [{\citenamefont {Will}\ \emph {et~al.}(2021)\citenamefont {Will},
  \citenamefont {Mathai}, \citenamefont {Huisman}, \citenamefont {Lohse},
  \citenamefont {Sun},\ and\ \citenamefont {Krug}}]{Will2021}%
  \BibitemOpen
  \bibfield  {author} {\bibinfo {author} {\bibfnamefont {J.~B.}\ \bibnamefont
  {Will}}, \bibinfo {author} {\bibfnamefont {V.}~\bibnamefont {Mathai}},
  \bibinfo {author} {\bibfnamefont {S.~G.}\ \bibnamefont {Huisman}}, \bibinfo
  {author} {\bibfnamefont {D.}~\bibnamefont {Lohse}}, \bibinfo {author}
  {\bibfnamefont {C.}~\bibnamefont {Sun}}, \ and\ \bibinfo {author}
  {\bibfnamefont {D.}~\bibnamefont {Krug}},\ }\href@noop {} {\bibfield
  {journal} {\bibinfo  {journal} {J. Fluid Mech}\ }\textbf {\bibinfo {volume}
  {912}} (\bibinfo {year} {2021})}\BibitemShut {NoStop}%
\bibitem [{\citenamefont {Verhille}\ and\ \citenamefont
  {Bartoli}(2016)}]{Verhille2016}%
  \BibitemOpen
  \bibfield  {author} {\bibinfo {author} {\bibfnamefont {G.}~\bibnamefont
  {Verhille}}\ and\ \bibinfo {author} {\bibfnamefont {A.}~\bibnamefont
  {Bartoli}},\ }\href@noop {} {\bibfield  {journal} {\bibinfo  {journal} {Exp.
  Fluids}\ }\textbf {\bibinfo {volume} {57}} (\bibinfo {year}
  {2016})}\BibitemShut {NoStop}%
\bibitem [{\citenamefont {Verhille}(2022)}]{Verhille2022}%
  \BibitemOpen
  \bibfield  {author} {\bibinfo {author} {\bibfnamefont {G.}~\bibnamefont
  {Verhille}},\ }\href@noop {} {\bibfield  {journal} {\bibinfo  {journal} {J.
  Fluid Mech.}\ }\textbf {\bibinfo {volume} {933}} (\bibinfo {year}
  {2022})}\BibitemShut {NoStop}%
\bibitem [{\citenamefont {Tinklenberg}, \citenamefont {Guala},\ and\
  \citenamefont {Coletti}(2023)}]{Tinklenberg2023}%
  \BibitemOpen
  \bibfield  {author} {\bibinfo {author} {\bibfnamefont {A.}~\bibnamefont
  {Tinklenberg}}, \bibinfo {author} {\bibfnamefont {M.}~\bibnamefont {Guala}},
  \ and\ \bibinfo {author} {\bibfnamefont {F.}~\bibnamefont {Coletti}},\
  }\href@noop {} {\bibfield  {journal} {\bibinfo  {journal} {J. Fluid Mech.}\
  }\textbf {\bibinfo {volume} {962}} (\bibinfo {year} {2023})}\BibitemShut
  {NoStop}%
\bibitem [{\citenamefont {Piumini}\ \emph {et~al.}(2024)\citenamefont
  {Piumini}, \citenamefont {Assen}, \citenamefont {Lohse},\ and\ \citenamefont
  {Verzicco}}]{Piumini2024}%
  \BibitemOpen
  \bibfield  {author} {\bibinfo {author} {\bibfnamefont {G.}~\bibnamefont
  {Piumini}}, \bibinfo {author} {\bibfnamefont {M.~P.~A.}\ \bibnamefont
  {Assen}}, \bibinfo {author} {\bibfnamefont {D.}~\bibnamefont {Lohse}}, \ and\
  \bibinfo {author} {\bibfnamefont {R.}~\bibnamefont {Verzicco}},\ }\href@noop
  {} {\bibfield  {journal} {\bibinfo  {journal} {J. Fluid Mech.}\ }\textbf
  {\bibinfo {volume} {995}} (\bibinfo {year} {2024})}\BibitemShut {NoStop}%
\bibitem [{\citenamefont {Toschi}\ and\ \citenamefont
  {Bodenschatz}(2009)}]{Toschi2009}%
  \BibitemOpen
  \bibfield  {author} {\bibinfo {author} {\bibfnamefont {F.}~\bibnamefont
  {Toschi}}\ and\ \bibinfo {author} {\bibfnamefont {E.}~\bibnamefont
  {Bodenschatz}},\ }\href@noop {} {\bibfield  {journal} {\bibinfo  {journal}
  {Annu. Rev. Fluid Mech.}\ }\textbf {\bibinfo {volume} {41}} (\bibinfo {year}
  {2009})}\BibitemShut {NoStop}%
\bibitem [{\citenamefont {Brandt}\ and\ \citenamefont
  {Coletti}(2022)}]{Brandt2022}%
  \BibitemOpen
  \bibfield  {author} {\bibinfo {author} {\bibfnamefont {L.}~\bibnamefont
  {Brandt}}\ and\ \bibinfo {author} {\bibfnamefont {F.}~\bibnamefont
  {Coletti}},\ }\href@noop {} {\bibfield  {journal} {\bibinfo  {journal} {Annu.
  Rev. Fluid Mech.}\ }\textbf {\bibinfo {volume} {54}} (\bibinfo {year}
  {2022})}\BibitemShut {NoStop}%
\bibitem [{\citenamefont {Harms}, \citenamefont {Brunton},\ and\ \citenamefont
  {McKeon}(2024)}]{Harms2024}%
  \BibitemOpen
  \bibfield  {author} {\bibinfo {author} {\bibfnamefont {T.}~\bibnamefont
  {Harms}}, \bibinfo {author} {\bibfnamefont {S.~L.}\ \bibnamefont {Brunton}},
  \ and\ \bibinfo {author} {\bibfnamefont {B.~J.}\ \bibnamefont {McKeon}},\
  }\href@noop {} {\bibfield  {journal} {\bibinfo  {journal} {ArXiv}\ }
  (\bibinfo {year} {2024})}\BibitemShut {NoStop}%
\bibitem [{\citenamefont {Cole}\ \emph {et~al.}(2016)\citenamefont {Cole},
  \citenamefont {Marcus}, \citenamefont {Parsa}, \citenamefont {Kramel},
  \citenamefont {Ni},\ and\ \citenamefont {Voth}}]{Cole2016}%
  \BibitemOpen
  \bibfield  {author} {\bibinfo {author} {\bibfnamefont {B.~C.}\ \bibnamefont
  {Cole}}, \bibinfo {author} {\bibfnamefont {G.~G.}\ \bibnamefont {Marcus}},
  \bibinfo {author} {\bibfnamefont {S.}~\bibnamefont {Parsa}}, \bibinfo
  {author} {\bibfnamefont {S.}~\bibnamefont {Kramel}}, \bibinfo {author}
  {\bibfnamefont {R.}~\bibnamefont {Ni}}, \ and\ \bibinfo {author}
  {\bibfnamefont {G.~A.}\ \bibnamefont {Voth}},\ }\href@noop {} {\bibfield
  {journal} {\bibinfo  {journal} {J. Vis. Exp.}\ }\textbf {\bibinfo {volume}
  {112}} (\bibinfo {year} {2016})}\BibitemShut {NoStop}%
\bibitem [{\citenamefont {Mathai}\ \emph {et~al.}(2016)\citenamefont {Mathai},
  \citenamefont {Neut}, \citenamefont {van~der Poel},\ and\ \citenamefont
  {Sun}}]{Mathai2016}%
  \BibitemOpen
  \bibfield  {author} {\bibinfo {author} {\bibfnamefont {V.}~\bibnamefont
  {Mathai}}, \bibinfo {author} {\bibfnamefont {M.~W.~M.}\ \bibnamefont {Neut}},
  \bibinfo {author} {\bibfnamefont {E.~P.}\ \bibnamefont {van~der Poel}}, \
  and\ \bibinfo {author} {\bibfnamefont {C.}~\bibnamefont {Sun}},\ }\href@noop
  {} {\bibfield  {journal} {\bibinfo  {journal} {Exp Fluids}\ }\textbf
  {\bibinfo {volume} {57}} (\bibinfo {year} {2016})}\BibitemShut {NoStop}%
\bibitem [{\citenamefont {Zimmermann}\ \emph {et~al.}(2012)\citenamefont
  {Zimmermann}, \citenamefont {Fiabane}, \citenamefont {Gasteuil},
  \citenamefont {Volk},\ and\ \citenamefont {Pinton}}]{Zimmermann2013}%
  \BibitemOpen
  \bibfield  {author} {\bibinfo {author} {\bibfnamefont {R.}~\bibnamefont
  {Zimmermann}}, \bibinfo {author} {\bibfnamefont {L.}~\bibnamefont {Fiabane}},
  \bibinfo {author} {\bibfnamefont {Y.}~\bibnamefont {Gasteuil}}, \bibinfo
  {author} {\bibfnamefont {R.}~\bibnamefont {Volk}}, \ and\ \bibinfo {author}
  {\bibfnamefont {J.-F.}\ \bibnamefont {Pinton}},\ }\href@noop {} {\bibfield
  {journal} {\bibinfo  {journal} {Phys. Scr.}\ } (\bibinfo {year}
  {2012})}\BibitemShut {NoStop}%
\bibitem [{\citenamefont {Ouellette}, \citenamefont {Xu},\ and\ \citenamefont
  {Bodenschatz}(2006)}]{Ouellette2006}%
  \BibitemOpen
  \bibfield  {author} {\bibinfo {author} {\bibfnamefont {N.~T.}\ \bibnamefont
  {Ouellette}}, \bibinfo {author} {\bibfnamefont {H.}~\bibnamefont {Xu}}, \
  and\ \bibinfo {author} {\bibfnamefont {E.}~\bibnamefont {Bodenschatz}},\
  }\href@noop {} {\bibfield  {journal} {\bibinfo  {journal} {Exp. Fluids}\
  }\textbf {\bibinfo {volume} {40}} (\bibinfo {year} {2006})}\BibitemShut
  {NoStop}%
\bibitem [{\citenamefont {Ibarra}, \citenamefont {Bartoli},\ and\ \citenamefont
  {Verhille}(2023)}]{Ibarra2023}%
  \BibitemOpen
  \bibfield  {author} {\bibinfo {author} {\bibfnamefont {E.}~\bibnamefont
  {Ibarra}}, \bibinfo {author} {\bibfnamefont {A.}~\bibnamefont {Bartoli}}, \
  and\ \bibinfo {author} {\bibfnamefont {G.}~\bibnamefont {Verhille}},\
  }\href@noop {} {\bibfield  {journal} {\bibinfo  {journal} {Exp. Fluids}\
  }\textbf {\bibinfo {volume} {64}} (\bibinfo {year} {2023})}\BibitemShut
  {NoStop}%
\bibitem [{\citenamefont {Parsa}\ \emph {et~al.}(2012)\citenamefont {Parsa},
  \citenamefont {Calzavarini}, \citenamefont {Toschi},\ and\ \citenamefont
  {Voth}}]{Parsa2012}%
  \BibitemOpen
  \bibfield  {author} {\bibinfo {author} {\bibfnamefont {S.}~\bibnamefont
  {Parsa}}, \bibinfo {author} {\bibfnamefont {E.}~\bibnamefont {Calzavarini}},
  \bibinfo {author} {\bibfnamefont {F.}~\bibnamefont {Toschi}}, \ and\ \bibinfo
  {author} {\bibfnamefont {G.~A.}\ \bibnamefont {Voth}},\ }\href@noop {}
  {\bibfield  {journal} {\bibinfo  {journal} {Phys. Rev. Lett.}\ }\textbf
  {\bibinfo {volume} {109}} (\bibinfo {year} {2012})}\BibitemShut {NoStop}%
\bibitem [{\citenamefont {Brizzolara}\ \emph {et~al.}(2021)\citenamefont
  {Brizzolara}, \citenamefont {Rosti}, \citenamefont {Olivieri}, \citenamefont
  {Brandt}, \citenamefont {Holzner},\ and\ \citenamefont
  {Mazzino}}]{Brizzolara2021}%
  \BibitemOpen
  \bibfield  {author} {\bibinfo {author} {\bibfnamefont {S.}~\bibnamefont
  {Brizzolara}}, \bibinfo {author} {\bibfnamefont {M.~E.}\ \bibnamefont
  {Rosti}}, \bibinfo {author} {\bibfnamefont {S.}~\bibnamefont {Olivieri}},
  \bibinfo {author} {\bibfnamefont {L.}~\bibnamefont {Brandt}}, \bibinfo
  {author} {\bibfnamefont {M.}~\bibnamefont {Holzner}}, \ and\ \bibinfo
  {author} {\bibfnamefont {A.}~\bibnamefont {Mazzino}},\ }\href@noop {}
  {\bibfield  {journal} {\bibinfo  {journal} {Phys. Rev. X}\ }\textbf {\bibinfo
  {volume} {11}} (\bibinfo {year} {2021})}\BibitemShut {NoStop}%
\bibitem [{\citenamefont {Bounoua}, \citenamefont {Bouchet},\ and\
  \citenamefont {Verhille}(2018)}]{Bounoua2018}%
  \BibitemOpen
  \bibfield  {author} {\bibinfo {author} {\bibfnamefont {S.}~\bibnamefont
  {Bounoua}}, \bibinfo {author} {\bibfnamefont {G.}~\bibnamefont {Bouchet}}, \
  and\ \bibinfo {author} {\bibfnamefont {G.}~\bibnamefont {Verhille}},\
  }\href@noop {} {\bibfield  {journal} {\bibinfo  {journal} {Phys. Rev. Lett.}\
  }\textbf {\bibinfo {volume} {121}} (\bibinfo {year} {2018})}\BibitemShut
  {NoStop}%
\bibitem [{\citenamefont {Shaik}\ \emph {et~al.}(2020)\citenamefont {Shaik},
  \citenamefont {Kuperman}, \citenamefont {Rinsky},\ and\ \citenamefont {van
  Hout}}]{Shaik2020}%
  \BibitemOpen
  \bibfield  {author} {\bibinfo {author} {\bibfnamefont {S.}~\bibnamefont
  {Shaik}}, \bibinfo {author} {\bibfnamefont {S.}~\bibnamefont {Kuperman}},
  \bibinfo {author} {\bibfnamefont {V.}~\bibnamefont {Rinsky}}, \ and\ \bibinfo
  {author} {\bibfnamefont {R.}~\bibnamefont {van Hout}},\ }\href@noop {}
  {\bibfield  {journal} {\bibinfo  {journal} {Phys. Rev. Fluids}\ }\textbf
  {\bibinfo {volume} {5}} (\bibinfo {year} {2020})}\BibitemShut {NoStop}%
\bibitem [{\citenamefont {Baker}\ and\ \citenamefont
  {Coletti}(2022)}]{Baker2022}%
  \BibitemOpen
  \bibfield  {author} {\bibinfo {author} {\bibfnamefont {L.~J.}\ \bibnamefont
  {Baker}}\ and\ \bibinfo {author} {\bibfnamefont {F.}~\bibnamefont
  {Coletti}},\ }\href@noop {} {\bibfield  {journal} {\bibinfo  {journal} {J.
  Fluid Mech}\ }\textbf {\bibinfo {volume} {943}} (\bibinfo {year}
  {2022})}\BibitemShut {NoStop}%
\bibitem [{\citenamefont {Zimmermann}\ \emph {et~al.}(2011)\citenamefont
  {Zimmermann}, \citenamefont {Gasteuil}, \citenamefont {Bourgoin},
  \citenamefont {Volk}, \citenamefont {Pumir},\ and\ \citenamefont
  {Pinton}}]{Zimmermann2011}%
  \BibitemOpen
  \bibfield  {author} {\bibinfo {author} {\bibfnamefont {R.}~\bibnamefont
  {Zimmermann}}, \bibinfo {author} {\bibfnamefont {Y.}~\bibnamefont
  {Gasteuil}}, \bibinfo {author} {\bibfnamefont {M.}~\bibnamefont {Bourgoin}},
  \bibinfo {author} {\bibfnamefont {R.}~\bibnamefont {Volk}}, \bibinfo {author}
  {\bibfnamefont {A.}~\bibnamefont {Pumir}}, \ and\ \bibinfo {author}
  {\bibfnamefont {J.}~\bibnamefont {Pinton}},\ }\href@noop {} {\bibfield
  {journal} {\bibinfo  {journal} {Rev. Sci. Instrum.}\ }\textbf {\bibinfo
  {volume} {82}} (\bibinfo {year} {2011})}\BibitemShut {NoStop}%
\bibitem [{\citenamefont {Niggel}\ \emph {et~al.}(2023)\citenamefont {Niggel},
  \citenamefont {Bailey}, \citenamefont {van Baalen}, \citenamefont {Zosso},\
  and\ \citenamefont {Isa}}]{Niggel2023}%
  \BibitemOpen
  \bibfield  {author} {\bibinfo {author} {\bibfnamefont {V.}~\bibnamefont
  {Niggel}}, \bibinfo {author} {\bibfnamefont {M.~R.}\ \bibnamefont {Bailey}},
  \bibinfo {author} {\bibfnamefont {C.}~\bibnamefont {van Baalen}}, \bibinfo
  {author} {\bibfnamefont {N.}~\bibnamefont {Zosso}}, \ and\ \bibinfo {author}
  {\bibfnamefont {L.}~\bibnamefont {Isa}},\ }\href@noop {} {\bibfield
  {journal} {\bibinfo  {journal} {Soft Matter}\ }\textbf {\bibinfo {volume}
  {19}} (\bibinfo {year} {2023})}\BibitemShut {NoStop}%
\bibitem [{\citenamefont {Marcus}\ \emph {et~al.}(2014)\citenamefont {Marcus},
  \citenamefont {Parsa}, \citenamefont {Kramel}, \citenamefont {Ni},\ and\
  \citenamefont {Voth}}]{Marcus2014}%
  \BibitemOpen
  \bibfield  {author} {\bibinfo {author} {\bibfnamefont {G.~G.}\ \bibnamefont
  {Marcus}}, \bibinfo {author} {\bibfnamefont {S.}~\bibnamefont {Parsa}},
  \bibinfo {author} {\bibfnamefont {S.}~\bibnamefont {Kramel}}, \bibinfo
  {author} {\bibfnamefont {R.}~\bibnamefont {Ni}}, \ and\ \bibinfo {author}
  {\bibfnamefont {G.~A.}\ \bibnamefont {Voth}},\ }\href@noop {} {\bibfield
  {journal} {\bibinfo  {journal} {New J. Phys.}\ }\textbf {\bibinfo {volume}
  {16}} (\bibinfo {year} {2014})}\BibitemShut {NoStop}%
\bibitem [{\citenamefont {Alipour}, \citenamefont {Paoli},\ and\ \citenamefont
  {S.~Ghaemi}(2021)}]{Alipour2021}%
  \BibitemOpen
  \bibfield  {author} {\bibinfo {author} {\bibfnamefont {M.}~\bibnamefont
  {Alipour}}, \bibinfo {author} {\bibfnamefont {M.~D.}\ \bibnamefont {Paoli}},
  \ and\ \bibinfo {author} {\bibfnamefont {A.~S.}\ \bibnamefont {S.~Ghaemi}},\
  }\href@noop {} {\bibfield  {journal} {\bibinfo  {journal} {J. Fluid Mech.}\
  }\textbf {\bibinfo {volume} {916}} (\bibinfo {year} {2021})}\BibitemShut
  {NoStop}%
\bibitem [{\citenamefont {Giurgiu}\ \emph {et~al.}(2024)\citenamefont
  {Giurgiu}, \citenamefont {Caridi}, \citenamefont {Paoli},\ and\ \citenamefont
  {Soldati}}]{Giurgiu2024}%
  \BibitemOpen
  \bibfield  {author} {\bibinfo {author} {\bibfnamefont {V.}~\bibnamefont
  {Giurgiu}}, \bibinfo {author} {\bibfnamefont {G.~C.~A.}\ \bibnamefont
  {Caridi}}, \bibinfo {author} {\bibfnamefont {M.~D.}\ \bibnamefont {Paoli}}, \
  and\ \bibinfo {author} {\bibfnamefont {A.}~\bibnamefont {Soldati}},\
  }\href@noop {} {\bibfield  {journal} {\bibinfo  {journal} {Phys. Rev. Lett.}\
  }\textbf {\bibinfo {volume} {133}} (\bibinfo {year} {2024})}\BibitemShut
  {NoStop}%
\bibitem [{\citenamefont {Fu}\ and\ \citenamefont {Liu}(2018)}]{Fu2018}%
  \BibitemOpen
  \bibfield  {author} {\bibinfo {author} {\bibfnamefont {Y.}~\bibnamefont
  {Fu}}\ and\ \bibinfo {author} {\bibfnamefont {Y.}~\bibnamefont {Liu}},\
  }\href@noop {} {\bibfield  {journal} {\bibinfo  {journal} {Meas. Sci.
  Technol.}\ }\textbf {\bibinfo {volume} {29}} (\bibinfo {year}
  {2018})}\BibitemShut {NoStop}%
\bibitem [{\citenamefont {Masuk}, \citenamefont {Salibindla},\ and\
  \citenamefont {Ni}(2019)}]{Masuk2019}%
  \BibitemOpen
  \bibfield  {author} {\bibinfo {author} {\bibfnamefont {A.~U.~M.}\
  \bibnamefont {Masuk}}, \bibinfo {author} {\bibfnamefont {A.}~\bibnamefont
  {Salibindla}}, \ and\ \bibinfo {author} {\bibfnamefont {R.}~\bibnamefont
  {Ni}},\ }\href@noop {} {\bibfield  {journal} {\bibinfo  {journal} {Int. J.
  Multiph. Flow}\ }\textbf {\bibinfo {volume} {120}} (\bibinfo {year}
  {2019})}\BibitemShut {NoStop}%
\bibitem [{\citenamefont {Tsai}(1987)}]{Tsai1987}%
  \BibitemOpen
  \bibfield  {author} {\bibinfo {author} {\bibfnamefont {R.}~\bibnamefont
  {Tsai}},\ }\href@noop {} {\bibfield  {journal} {\bibinfo  {journal} {IEEE J.
  Robot. Autom.}\ }\textbf {\bibinfo {volume} {3}} (\bibinfo {year}
  {1987})}\BibitemShut {NoStop}%
\bibitem [{\citenamefont {Bourgoin}\ and\ \citenamefont
  {Huisman}(2020)}]{Bourgoin2020}%
  \BibitemOpen
  \bibfield  {author} {\bibinfo {author} {\bibfnamefont {M.}~\bibnamefont
  {Bourgoin}}\ and\ \bibinfo {author} {\bibfnamefont {S.~G.}\ \bibnamefont
  {Huisman}},\ }\href@noop {} {\bibfield  {journal} {\bibinfo  {journal} {Rev.
  Sci. Instrum.}\ }\textbf {\bibinfo {volume} {91}} (\bibinfo {year}
  {2020})}\BibitemShut {NoStop}%
\bibitem [{\citenamefont {Giusto}\ \emph {et~al.}(2024)\citenamefont {Giusto},
  \citenamefont {Bergougnoux}, \citenamefont {Marchioli},\ and\ \citenamefont
  {Guazzelli}}]{diGiusto2024}%
  \BibitemOpen
  \bibfield  {author} {\bibinfo {author} {\bibfnamefont {D.~D.}\ \bibnamefont
  {Giusto}}, \bibinfo {author} {\bibfnamefont {L.}~\bibnamefont {Bergougnoux}},
  \bibinfo {author} {\bibfnamefont {C.}~\bibnamefont {Marchioli}}, \ and\
  \bibinfo {author} {\bibfnamefont {E.}~\bibnamefont {Guazzelli}},\ }\href@noop
  {} {\bibfield  {journal} {\bibinfo  {journal} {J. Fluid Mech}\ }\textbf
  {\bibinfo {volume} {979}} (\bibinfo {year} {2024})}\BibitemShut {NoStop}%
\bibitem [{\citenamefont {Gonzalez}\ and\ \citenamefont
  {Woods}(2002)}]{DigitalImageProcessing}%
  \BibitemOpen
  \bibfield  {author} {\bibinfo {author} {\bibfnamefont {R.~C.}\ \bibnamefont
  {Gonzalez}}\ and\ \bibinfo {author} {\bibfnamefont {R.~E.}\ \bibnamefont
  {Woods}},\ }\href@noop {} {\emph {\bibinfo {title} {Digital Image
  Processing}}},\ \bibinfo {edition} {2nd}\ ed.\ (\bibinfo  {publisher}
  {Prentice-Hall},\ \bibinfo {year} {2002})\BibitemShut {NoStop}%
\bibitem [{\citenamefont {Rother}, \citenamefont {Kolmogorov},\ and\
  \citenamefont {Blake}(2004)}]{Rother2004}%
  \BibitemOpen
  \bibfield  {author} {\bibinfo {author} {\bibfnamefont {C.}~\bibnamefont
  {Rother}}, \bibinfo {author} {\bibfnamefont {V.}~\bibnamefont {Kolmogorov}},
  \ and\ \bibinfo {author} {\bibfnamefont {A.}~\bibnamefont {Blake}},\
  }\href@noop {} {\bibfield  {journal} {\bibinfo  {journal} {ACM Trans.
  Graph.}\ }\textbf {\bibinfo {volume} {23}} (\bibinfo {year}
  {2004})}\BibitemShut {NoStop}%
\bibitem [{\citenamefont {Bruijne}\ and\ \citenamefont
  {Nielsen}(2004)}]{DeBruijne2004}%
  \BibitemOpen
  \bibfield  {author} {\bibinfo {author} {\bibfnamefont {M.~D.}\ \bibnamefont
  {Bruijne}}\ and\ \bibinfo {author} {\bibfnamefont {M.}~\bibnamefont
  {Nielsen}},\ }\href@noop {} {\bibfield  {journal} {\bibinfo  {journal}
  {Medical Image Computing and Computer-Assisted Intervention–MICCAI 2004:
  7th International Conference}\ } (\bibinfo {year} {2004})}\BibitemShut
  {NoStop}%
\bibitem [{\citenamefont {Nelder}\ and\ \citenamefont
  {Mead}(1965)}]{Nelder1965}%
  \BibitemOpen
  \bibfield  {author} {\bibinfo {author} {\bibfnamefont {J.~A.}\ \bibnamefont
  {Nelder}}\ and\ \bibinfo {author} {\bibfnamefont {R.}~\bibnamefont {Mead}},\
  }\href@noop {} {\bibfield  {journal} {\bibinfo  {journal} {Comput. J}\ }
  (\bibinfo {year} {1965})}\BibitemShut {NoStop}%
\bibitem [{\citenamefont {Shen}\ \emph {et~al.}(2000)\citenamefont {Shen},
  \citenamefont {Song}, \citenamefont {Iguchi},\ and\ \citenamefont
  {Yamamoto}}]{Shen2000}%
  \BibitemOpen
  \bibfield  {author} {\bibinfo {author} {\bibfnamefont {L.}~\bibnamefont
  {Shen}}, \bibinfo {author} {\bibfnamefont {X.}~\bibnamefont {Song}}, \bibinfo
  {author} {\bibfnamefont {M.}~\bibnamefont {Iguchi}}, \ and\ \bibinfo {author}
  {\bibfnamefont {F.}~\bibnamefont {Yamamoto}},\ }\href@noop {} {\bibfield
  {journal} {\bibinfo  {journal} {Pattern Recognition Letters}\ }\textbf
  {\bibinfo {volume} {21}} (\bibinfo {year} {2000})}\BibitemShut {NoStop}%
\bibitem [{\citenamefont {Neoptolemou}\ \emph {et~al.}(2022)\citenamefont
  {Neoptolemou}, \citenamefont {Goyal}, \citenamefont {Cruz-Cabeza},
  \citenamefont {Kiss}, \citenamefont {Milne},\ and\ \citenamefont
  {Vetter}}]{Neoptolemou2022}%
  \BibitemOpen
  \bibfield  {author} {\bibinfo {author} {\bibfnamefont {P.}~\bibnamefont
  {Neoptolemou}}, \bibinfo {author} {\bibfnamefont {N.}~\bibnamefont {Goyal}},
  \bibinfo {author} {\bibfnamefont {A.~J.}\ \bibnamefont {Cruz-Cabeza}},
  \bibinfo {author} {\bibfnamefont {A.~A.}\ \bibnamefont {Kiss}}, \bibinfo
  {author} {\bibfnamefont {D.~J.}\ \bibnamefont {Milne}}, \ and\ \bibinfo
  {author} {\bibfnamefont {T.}~\bibnamefont {Vetter}},\ }\href@noop {}
  {\bibfield  {journal} {\bibinfo  {journal} {Powder Technology}\ }\textbf
  {\bibinfo {volume} {399}} (\bibinfo {year} {2022})}\BibitemShut {NoStop}%
\bibitem [{\citenamefont {Puzyrev}\ \emph {et~al.}(2020)\citenamefont
  {Puzyrev}, \citenamefont {Harth}, \citenamefont {Trittel},\ and\
  \citenamefont {Stannarius}}]{Puzyrev2020}%
  \BibitemOpen
  \bibfield  {author} {\bibinfo {author} {\bibfnamefont {D.}~\bibnamefont
  {Puzyrev}}, \bibinfo {author} {\bibfnamefont {K.}~\bibnamefont {Harth}},
  \bibinfo {author} {\bibfnamefont {T.}~\bibnamefont {Trittel}}, \ and\
  \bibinfo {author} {\bibfnamefont {R.}~\bibnamefont {Stannarius}},\
  }\href@noop {} {\bibfield  {journal} {\bibinfo  {journal} {Microgravity Sci.
  Technol.}\ }\textbf {\bibinfo {volume} {32}} (\bibinfo {year}
  {2020})}\BibitemShut {NoStop}%
\bibitem [{\citenamefont {Oehmke}\ \emph {et~al.}(2021)\citenamefont {Oehmke},
  \citenamefont {Bordoloi}, \citenamefont {Variano},\ and\ \citenamefont
  {Verhille}}]{Oehmke2021}%
  \BibitemOpen
  \bibfield  {author} {\bibinfo {author} {\bibfnamefont {T.~B.}\ \bibnamefont
  {Oehmke}}, \bibinfo {author} {\bibfnamefont {A.~D.}\ \bibnamefont
  {Bordoloi}}, \bibinfo {author} {\bibfnamefont {E.}~\bibnamefont {Variano}}, \
  and\ \bibinfo {author} {\bibfnamefont {G.}~\bibnamefont {Verhille}},\
  }\href@noop {} {\bibfield  {journal} {\bibinfo  {journal} {Phys. Rev.
  Fluids}\ }\textbf {\bibinfo {volume} {6}} (\bibinfo {year}
  {2021})}\BibitemShut {NoStop}%
\bibitem [{\citenamefont {van Gils}\ \emph {et~al.}(2011)\citenamefont {van
  Gils}, \citenamefont {Bruggert}, \citenamefont {Lathrop}, \citenamefont
  {Sun},\ and\ \citenamefont {Lohse}}]{vanGils2011}%
  \BibitemOpen
  \bibfield  {author} {\bibinfo {author} {\bibfnamefont {D.~P.~M.}\
  \bibnamefont {van Gils}}, \bibinfo {author} {\bibfnamefont {G.~W.~H.}\
  \bibnamefont {Bruggert}}, \bibinfo {author} {\bibfnamefont {D.~P.}\
  \bibnamefont {Lathrop}}, \bibinfo {author} {\bibfnamefont {C.}~\bibnamefont
  {Sun}}, \ and\ \bibinfo {author} {\bibfnamefont {D.}~\bibnamefont {Lohse}},\
  }\href@noop {} {\bibfield  {journal} {\bibinfo  {journal} {Rev. Sci.
  Instrum.}\ }\textbf {\bibinfo {volume} {82}} (\bibinfo {year}
  {2011})}\BibitemShut {NoStop}%
\bibitem [{\citenamefont {Huisman}\ \emph {et~al.}(2015)\citenamefont
  {Huisman}, \citenamefont {van~der Veen}, \citenamefont {Bruggert},
  \citenamefont {Lohse},\ and\ \citenamefont {Sun.}}]{Huisman2015}%
  \BibitemOpen
  \bibfield  {author} {\bibinfo {author} {\bibfnamefont {S.~G.}\ \bibnamefont
  {Huisman}}, \bibinfo {author} {\bibfnamefont {R.~C.~A.}\ \bibnamefont
  {van~der Veen}}, \bibinfo {author} {\bibfnamefont {G.~W.~H.}\ \bibnamefont
  {Bruggert}}, \bibinfo {author} {\bibfnamefont {D.}~\bibnamefont {Lohse}}, \
  and\ \bibinfo {author} {\bibfnamefont {C.}~\bibnamefont {Sun.}},\ }\href@noop
  {} {\bibfield  {journal} {\bibinfo  {journal} {Rev. Sci. Instrum.}\ }\textbf
  {\bibinfo {volume} {86}} (\bibinfo {year} {2015})}\BibitemShut {NoStop}%
\bibitem [{\citenamefont {van~der Veen}\ \emph {et~al.}(2016)\citenamefont
  {van~der Veen}, \citenamefont {Huisman}, \citenamefont {Dung}, \citenamefont
  {Tang}, \citenamefont {Sun},\ and\ \citenamefont {Lohse}}]{vanderVeen2016}%
  \BibitemOpen
  \bibfield  {author} {\bibinfo {author} {\bibfnamefont {R.~C.~A.}\
  \bibnamefont {van~der Veen}}, \bibinfo {author} {\bibfnamefont {S.~G.}\
  \bibnamefont {Huisman}}, \bibinfo {author} {\bibfnamefont {O.~Y.}\
  \bibnamefont {Dung}}, \bibinfo {author} {\bibfnamefont {H.~L.}\ \bibnamefont
  {Tang}}, \bibinfo {author} {\bibfnamefont {C.}~\bibnamefont {Sun}}, \ and\
  \bibinfo {author} {\bibfnamefont {D.}~\bibnamefont {Lohse}},\ }\href@noop {}
  {\bibfield  {journal} {\bibinfo  {journal} {Phys. Rev. Fluids}\ }\textbf
  {\bibinfo {volume} {1}} (\bibinfo {year} {2016})}\BibitemShut {NoStop}%
\bibitem [{\citenamefont {Porta}, \citenamefont {Voth},\ and\ \citenamefont
  {Crawford}(2001)}]{LaPorta2001}%
  \BibitemOpen
  \bibfield  {author} {\bibinfo {author} {\bibfnamefont {A.~L.}\ \bibnamefont
  {Porta}}, \bibinfo {author} {\bibfnamefont {G.~A.}\ \bibnamefont {Voth}}, \
  and\ \bibinfo {author} {\bibfnamefont {A.}~\bibnamefont {Crawford}},\
  }\href@noop {} {\bibfield  {journal} {\bibinfo  {journal} {Nature}\ }\textbf
  {\bibinfo {volume} {40}} (\bibinfo {year} {2001})}\BibitemShut {NoStop}%
\bibitem [{\citenamefont {Calzavarini}\ \emph {et~al.}(2009)\citenamefont
  {Calzavarini}, \citenamefont {Volk}, \citenamefont {Bourgoin}, \citenamefont
  {Lévêque}, \citenamefont {Pinton},\ and\ \citenamefont
  {Toschi}}]{Calzavarini2009}%
  \BibitemOpen
  \bibfield  {author} {\bibinfo {author} {\bibfnamefont {E.}~\bibnamefont
  {Calzavarini}}, \bibinfo {author} {\bibfnamefont {R.}~\bibnamefont {Volk}},
  \bibinfo {author} {\bibfnamefont {M.}~\bibnamefont {Bourgoin}}, \bibinfo
  {author} {\bibfnamefont {E.}~\bibnamefont {Lévêque}}, \bibinfo {author}
  {\bibfnamefont {J.-F.}\ \bibnamefont {Pinton}}, \ and\ \bibinfo {author}
  {\bibfnamefont {F.}~\bibnamefont {Toschi}},\ }\href@noop {} {\bibfield
  {journal} {\bibinfo  {journal} {J. Fluid Mech.}\ }\textbf {\bibinfo {volume}
  {630}} (\bibinfo {year} {2009})}\BibitemShut {NoStop}%
\bibitem [{\citenamefont {Klein}\ \emph {et~al.}(2012)\citenamefont {Klein},
  \citenamefont {Gibert}, \citenamefont {Bérut},\ and\ \citenamefont
  {Bodenschatz}}]{Klein2013}%
  \BibitemOpen
  \bibfield  {author} {\bibinfo {author} {\bibfnamefont {S.}~\bibnamefont
  {Klein}}, \bibinfo {author} {\bibfnamefont {M.}~\bibnamefont {Gibert}},
  \bibinfo {author} {\bibfnamefont {A.}~\bibnamefont {Bérut}}, \ and\ \bibinfo
  {author} {\bibfnamefont {E.}~\bibnamefont {Bodenschatz}},\ }\href@noop {}
  {\bibfield  {journal} {\bibinfo  {journal} {Meas. Sci. Technol.}\ }\textbf
  {\bibinfo {volume} {24}} (\bibinfo {year} {2012})}\BibitemShut {NoStop}%
\bibitem [{\citenamefont {Zimmermann}\ \emph {et~al.}(2010)\citenamefont
  {Zimmermann}, \citenamefont {Xu}, \citenamefont {Gasteuil}, \citenamefont
  {Bourgoin}, \citenamefont {Volk}, \citenamefont {Pinton},\ and\ \citenamefont
  {Bodenschatz}}]{Zimmermann2010}%
  \BibitemOpen
  \bibfield  {author} {\bibinfo {author} {\bibfnamefont {R.}~\bibnamefont
  {Zimmermann}}, \bibinfo {author} {\bibfnamefont {H.}~\bibnamefont {Xu}},
  \bibinfo {author} {\bibfnamefont {Y.}~\bibnamefont {Gasteuil}}, \bibinfo
  {author} {\bibfnamefont {M.}~\bibnamefont {Bourgoin}}, \bibinfo {author}
  {\bibfnamefont {R.}~\bibnamefont {Volk}}, \bibinfo {author} {\bibfnamefont
  {J.-F.}\ \bibnamefont {Pinton}}, \ and\ \bibinfo {author} {\bibfnamefont
  {E.}~\bibnamefont {Bodenschatz}},\ }\href@noop {} {\bibfield  {journal}
  {\bibinfo  {journal} {Rev. Sci. Instrum.}\ }\textbf {\bibinfo {volume} {81}}
  (\bibinfo {year} {2010})}\BibitemShut {NoStop}%
\end{thebibliography}
\end{document}